%% file: main.tex
\documentclass[10pt,twocolumn,letterpaper]{article}

\usepackage{iccv}
\usepackage{times}
\usepackage{epsfig}
\usepackage{graphicx}
\usepackage{amsmath}
\usepackage{amssymb}
\usepackage{amsfonts}
\usepackage{multirow}
\usepackage{booktabs}
 \usepackage{booktabs}
 \usepackage{multirow}
 \usepackage{enumitem}

\newcommand{\cusp}{CUSP}

\newcommand{\colwidth}{0.24\linewidth}
\newcommand{\gap}{@{\hspace{0.5mm}}}

\usepackage[pagebackref=true,breaklinks=true,letterpaper=true,colorlinks,bookmarks=false]{hyperref}
\iccvfinalcopy 

\def\iccvPaperID{8725} 

\ificcvfinal\fi

\begin{document}

\title{Uncertainty Surrogates for Deep Learning}

\author{Radhakrishna Achanta and Natasa Tagasovska\\
Swiss Data Science Center\\
EPFL and ETH Zurich \\
Switzerland\\
{\tt\small radhakrishna.achanta, natasa.tagasovska@epfl.ch}
}

\maketitle
\ificcvfinal\thispagestyle{empty}\fi

\input{latex/10_abstract.tex}


\input{latex/20_intro.tex}
\input{latex/30_background.tex}
\input{latex/40_method.tex}
\input{latex/50_experiments.tex}
\input{latex/60_conclusion.tex}




{\small
\bibliographystyle{ieee_fullname}
\bibliography{egbib}
}

\end{document}

%% file: latex/10_abstract.tex
\begin{abstract}
In this paper we introduce a novel way of estimating prediction uncertainty in deep networks through the use of uncertainty surrogates. These surrogates are features of the penultimate layer of a deep network that are forced to match predefined patterns. The patterns themselves can be, among other possibilities, a known visual symbol. We show how our approach can be used for estimating uncertainty in prediction and out-of-distribution detection. Additionally, the surrogates allow for interpretability of the ability of the deep network to learn and at the same time lend robustness against adversarial attacks. Despite its simplicity, our approach is superior to the state-of-the-art approaches on standard metrics as well as computational efficiency and ease of implementation. A wide range of experiments are performed on standard datasets to prove the efficacy of our approach.
\end{abstract}

%% file: latex/20_intro.tex
\section{Introduction}
Deep learning has made large strides in recent times, providing breakthroughs in several computer vision and machine learning problems. However, deep learning models remain unable to assess their confidence when performing under unforeseen situations.
As argued by Begoli \etal~\cite{begoli2019need}, uncertainty quantification is a problem of paramount importance when deploying machine learning models in sensitive contexts such as information security~\cite{smith2011information}, engineering~\cite{wen2003uncertainty}, transportation~\cite{zhu2017deep}, or medicine~\cite{begoli2019need}.

\begin{figure}
	\includegraphics[width=1.0\columnwidth]{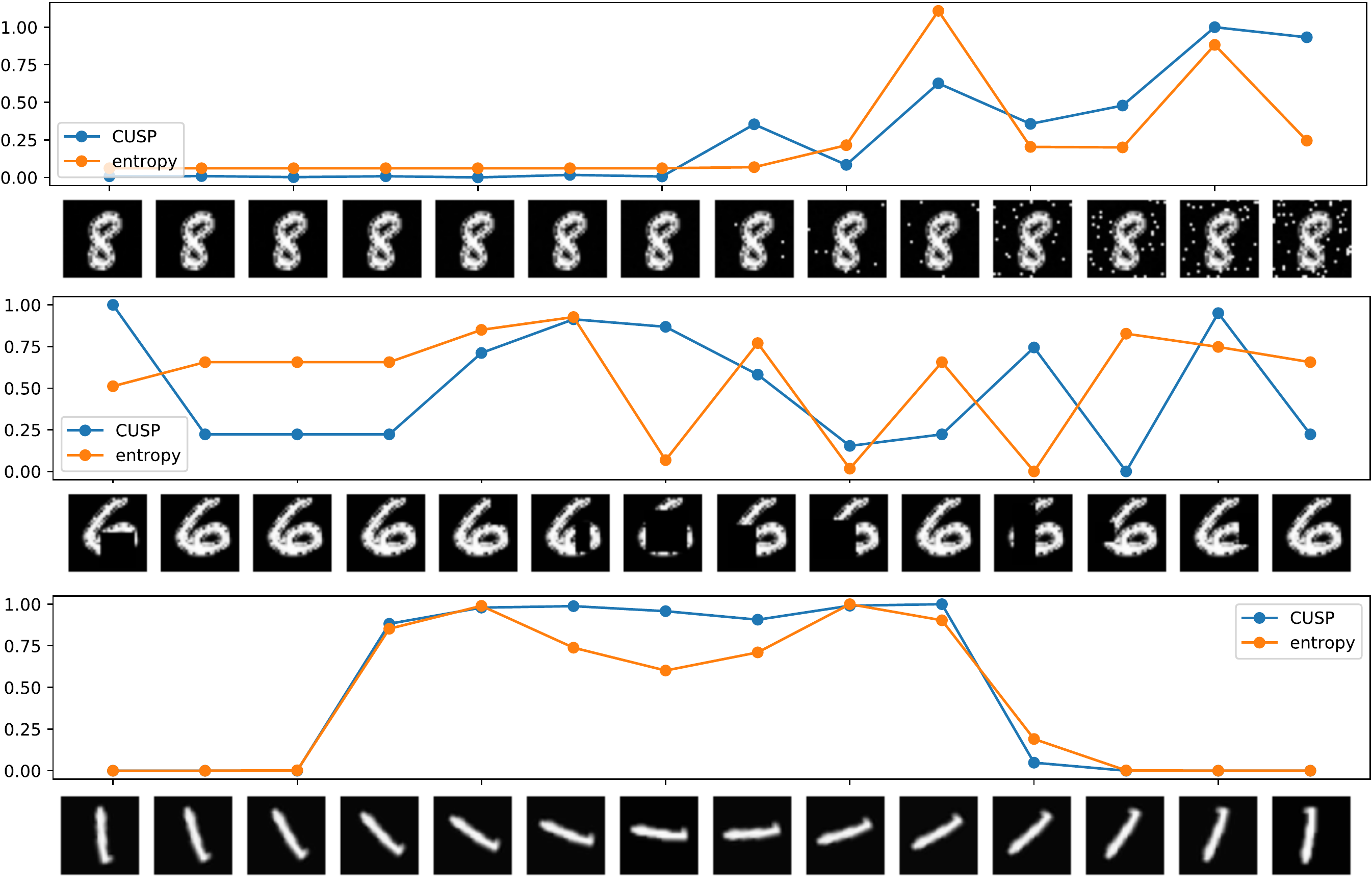}
	\caption{For three different corruptions of the original image, namely noise addition, random cropping, and rotation, compared to a competing technique~\cite{wang2014new}, our approach shows the uncertainty in prediction that better reflects the qualitative deterioration of the original image.}
\end{figure}
\label{fig:motivating_example}

Uncertainty estimation can be used to assess risk or decide when to abstain from using the prediction. Such abstention is a reasonable strategy to deal with anomalies~\cite{chandola2009anomaly}, outliers~\cite{hodge2004survey}, out-of-distribution examples~\cite{shafaei2018does}, to detect and defend against adversaries~\cite{intriguing}, or to delegate high-risk predictions to humans~\cite{cortes2016learning, geifman2017selective, chen2018confidence}. Knowing uncertainty is the backbone of active learning~\cite{settles2014active}, when it comes to optimally choosing examples a human should annotate. Importantly, uncertainty quantification takes us closer to model interpretability~\cite{alvarez2017causal}.

In this paper, we address this outstanding problem, namely, quantification of the degree of uncertainty of the inference done by a deep network, with a simple yet powerful method. Our approach consists of assigning, to each class, a fixed binary pattern, often visually meaningful (see \autoref{fig:intro_visual}), and forcing the activations of the penultimate layer to resemble it using an additional reconstruction loss.

Training with this additional loss turns the penultimate layer activations into proxies for estimating the uncertainty of classification results. Our method is hence termed \emph{Classification Uncertainty from Surrogate Prediction} or \cusp, in short. Additionally, in the rest of the paper we refer to the predefined patterns used for training uncertainty surrogates as \emph{surrogate patterns}.


\input{./latex/22_fig_cifar10strip.tex}
Despite its apparent simplicity, in one fell swoop CUSP addresses several outstanding issues related to prediction uncertainty, which are otherwise separately addressed by existing works. Our contributions are as follows:
\begin{itemize}[noitemsep]
    \item Uncertainty surrogates help quantify prediction uncertainty through reliable, yet computationally efficient measures.
    \item We show through experiments how anomalies, outliers, and corrupted examples, i.e out-of-distribution examples, can be detected with \cusp.
    \item The regularization introduced by {\cusp} during training improves robustness against adversarial attacks.
    \item Due to the visual nature of the surrogates, we show that a secondary detector can be trained to quantify the uncertitude of the classifier.
    \item {\cusp} allows for visual inspection of the training progress as well the test results, which along with quantified uncertainty values provides interpretability.
\end{itemize}

We run several experiments over standard datasets to show the efficacy of our approach and its superiority as compared to the state-of-the-art methods that usually are able to address only a subset of the issues we tackle.

The rest of the paper is organized as follows: we present how uncertainty is defined and categorized in literature along with the work in \autoref{sec:background}. This positions our method CUSP, which we introduce in detail in \autoref{sec:method}. In \autoref{sec:experiments} we provide extensive empirical results for the applicability and trustworthiness of CUSP. Finally, we discuss and conclude in \autoref{sec:conclusion}. 


%% file: latex/22_fig_cifar10strip.tex
\renewcommand{\colwidth}{ 0.06\linewidth}
\renewcommand{\gap}{@{\hspace{0.5mm}}}

\begin{figure*}
\centering

\begin{tabular}{c\gap c\gap c\gap c\gap c\gap c\gap c\gap c\gap c\gap c\gap c\gap c\gap c\gap c\gap c\gap}

    \includegraphics[width=\colwidth]{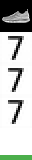}&
    \includegraphics[width=\colwidth]{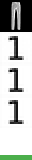}&
    \includegraphics[width=\colwidth]{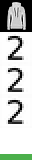}&
    \includegraphics[width=\colwidth]{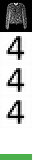}&
    \includegraphics[width=\colwidth]{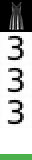}&
    \includegraphics[width=\colwidth]{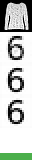}&
    \includegraphics[width=\colwidth]{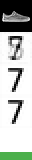}&
    \includegraphics[width=\colwidth]{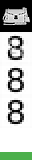}&
    \includegraphics[width=\colwidth]{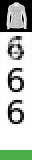}&
    \includegraphics[width=\colwidth]{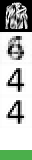}&
    \includegraphics[width=\colwidth]{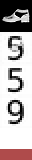}&
    \includegraphics[width=\colwidth]{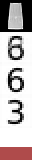}&
    \includegraphics[width=\colwidth]{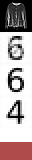}&
    \includegraphics[width=\colwidth]{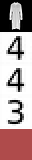}&
    \includegraphics[width=\colwidth]{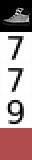}\\
    
    0.094 &
    0.097 &
    0.117 &
    0.119 &
    0.121 &
    0.145 &
    0.154 &
    0.166 &
    0.198 &
    0.204 &
    0.213 &
    0.254 &
    0.368 &
    0.612 &
    0.64 \\
&&&&&&&&&&&&&&\\
\includegraphics[width=\colwidth]{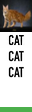}&
\includegraphics[width=\colwidth]{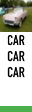}&
\includegraphics[width=\colwidth]{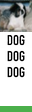}&
\includegraphics[width=\colwidth]{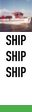}&
\includegraphics[width=\colwidth]{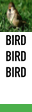}&
\includegraphics[width=\colwidth]{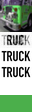}&
\includegraphics[width=\colwidth]{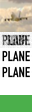}&
\includegraphics[width=\colwidth]{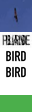}&
\includegraphics[width=\colwidth]{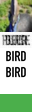}&
\includegraphics[width=\colwidth]{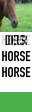}&
\includegraphics[width=\colwidth]{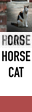}&
\includegraphics[width=\colwidth]{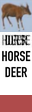}&
\includegraphics[width=\colwidth]{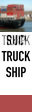}&
\includegraphics[width=\colwidth]{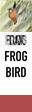}&
\includegraphics[width=\colwidth]{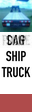}\\

0.108 &
0.127 &
0.128 &
0.146 &
0.158 &
0.337 &
0.344 &
0.347 &
0.354 &
0.393 &
0.328 &
0.337 &
0.341 &
0.348 &
0.362 \\
&&&&&&&&&&&&&&\\

\end{tabular}
\caption{Visualizations using our approach correspondng to FashionMNIST and CIFAR-10 datasets. The FashionMNIST images are trained with ten $16\times 16$ digit images, while the CIFAR-10 images are trained with ten $16\times 32$ text images containing the corresponding class name. In each of the two sets, the rows from top contain: the original images from the CIFAR-10 dataset, our uncertainty surrogates, the pattern corresponding to the predicted label, the pattern corresponding to the true label, the uncertainty score as a colored bar (green for correctly predicted and red for incorrectly predicted examples), and the numerical value of the uncertainty score. In the columns from left to right: first five images show low uncertainty and are correctly classified, the second five images show high uncertainty but are still correctly classified, and last five images show high uncertainty and are wrongly classified. As can be seen the uncertainty computed through our surrogates is a good indicator of the quality of the decision.}
\label{fig:intro_visual}
\end{figure*}

%% file: latex/30_background.tex
\section{Background}
\label{sec:background}



The lack of confidence estimates being an essential drawback for the advancements of deep learning at production level, has inspired lot of recent works addressing this challenge~\cite{hernandez2015probabilistic, Gal2016Uncertainty, lakshminarayanan2017simple, kendall2017uncertainties, tagasovska2018single}.

Research in the domain of uncertainty estimation could be categorized in multiple ways: based on \emph{implementation technique}: augmenting the network architecture, the loss function, or the data itself;  based on the \emph{machine learning task}: classification, regression, reinforcement learning etc;  or based on the \emph{source of uncertainty}: aleatoric or epistemic.

\begin{figure}
    \centering
    \includegraphics[width=1.0\columnwidth]{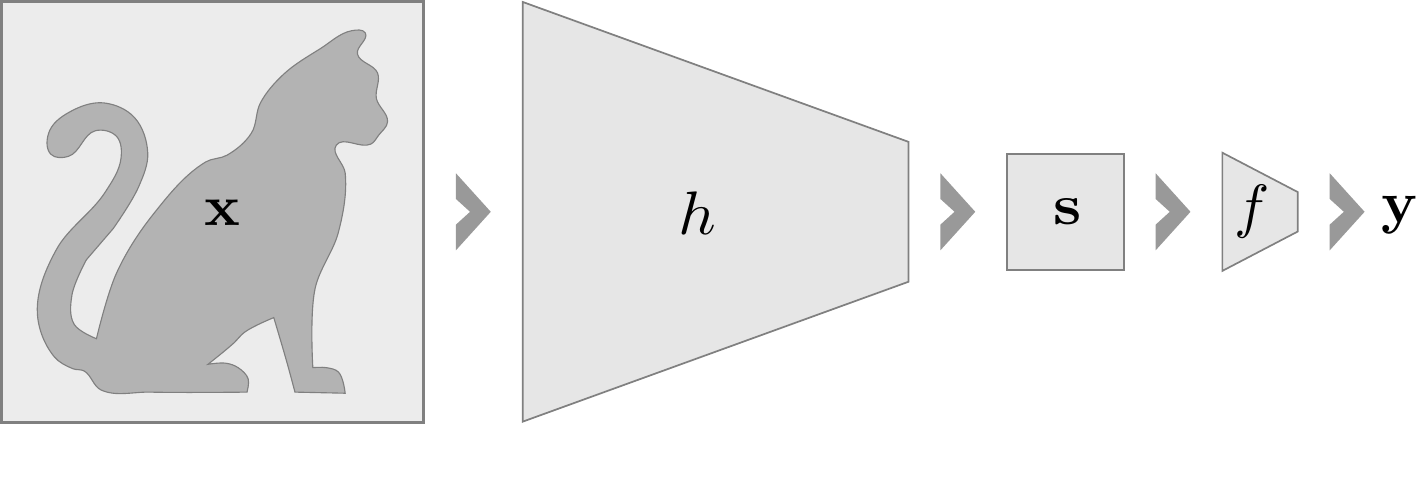}
    \caption{A deep network $\mathcal{F}$ generates a surrogate feature layer $\mathbf{s}$ from an input image $\mathbf{x}$, which is then used for predicting $\mathbf{y}$ as well as for assessing the uncertainty of the prediction $\mathbf{y}$ by comparing the pre-determined pattern that $\mathbf{s}$ is expected to resemble.}
    \label{fig:schema}
\end{figure}

For the sake of computational efficiency, one of the most popular approaches is using Monte-Carlo dropout~\cite{Gal2016Uncertainty}, that provides an estimate for the predictive uncertainty by aggregating results from multiple dropout masks. Another method favoured for its intuitive approach is ensembling, wherein multiple models are trained and the inherent variance between the prediction of the models is used as a proxy for uncertainty~\cite{lakshminarayanan2017simple}. Both of these methods require custom loss functions and/or modifications to the architecture.

Several variants of these initial models were introduced as an attempt to either consider the two sources of uncertainty~\cite{kendall2017uncertainties} for MC dropout, or to reduce the computational overhead~\cite{havasi2020training, tran2020hydra, wen2020batchensemble}, or to combine dropout with ensembles~\cite{durasov2020masksembles}.


Since it is quite challenging to develop methods targeting a specific ML application or architecture, several existing approaches focus on the task at hand. Some focus on classification~\cite{sensoy2018evidential, hein2019relu, hendrycks2019augmix, dusenberry2020efficient, malinin2018prior}, while others address regression tasks~\cite{pearce2018high, brando2019modelling, qiu2019quantifying} with the goal of obtaining high quality prediction intervals, say, by allowing for non-Gaussian, non i.i.d noise in the data.

A convenient choice for estimating uncertainty is to leverage the feature representations at the final layer to distinguish between in-distribution and out-of-distribution samples~\cite{tagasovska2018single, burda2018exploration,odin}. Tagasovska \etal~\cite{tagasovska2018single} show that while there are novel methods being introduced, baseline methods such as the entropy of the softmax layer and its variants already provide hard-to-beat uncertainty estimates for out-of-distribution detection. Our work, {\cusp} relies on the pen-utlimate layer of networks, which can be related in spirit to these simple yet seasoned methods.

Formally, uncertainty in machine learning is defined with regards to its source.
The term \emph{aleatoric} (from the latin word ``alea" meaning randomness) uncertainty relates to the inherent stohasticity in the data, i.e. noise, which is irreducible, but, should be accounted for in a model. 
The term \emph{epistemic} (from the Greek word ``episteme" i.e. knowledge) uncertainty relates to incomplete knowledge, whether that emanates from the lack of observations in certain input regions, the presence of unmeasured variables, or the propagation of errors from preceding operations for computing the target values.

In~\autoref{tab:overview}, we distinguish as well as summarize the cases one should consider when deciding on which type of uncertainty is most relevant for their application. For example, guarding against adversarial attacks is related to data noise in the input or target variables, whereas anomaly detection or distribution shift is related to the epistemic uncertainty of the model. 

\begin{table}[t]
	
		\resizebox{0.5\textwidth}{!}{
\begin{tabular}{lcc}
	\hline
	\multicolumn{1}{c}{} & \textbf{Aleatoric}                                                                       & \textbf{Epistemic}                                                              \\ \hline
	\textbf{Inputs}     $X$ & \begin{tabular}[c]{@{}c@{}}noisy measurements;\\ faulty equipment\end{tabular}           & \begin{tabular}[c]{@{}c@{}}unobs. variables\\ unobs. input regions\end{tabular} \\ \cline{2-3} 
	\textbf{Targets}   $Y$  & \begin{tabular}[c]{@{}c@{}} $y$  derived from noisy $x$\\ disagreed class labels\end{tabular} & \begin{tabular}[c]{@{}c@{}}wrong numerical \\ /labeling model\end{tabular}         \\ \hline
	& & \\
\end{tabular}
}
\caption{Overview of the categories of data-related uncertainty machine learning. }
\label{tab:overview}
\end{table}

Owing to methodological limitations or developing approaches with a certain application in mind, most of the existing methods determining uncertainty address only one or two of the cells in~\autoref{tab:overview}.  
In this work we are interested computing an uncertainty estimate for deep learning that is simultaneously sensitive to both aleatoric and epistemic uncertainty for the task of classification.

In~\autoref{fig:motivating_example} we show how uncertainties obtained from our approach {\cusp} relate to three different situations. In the first row, we notice how the uncertainty score increases with an additive random noise. In the second row we erase parts of the image at random and again we notice the surrogates are very sensitive to the missing patches. Both of these corruptions are related to the aleatoric source of noise.

If, on the other hand, we rotate a given image from 0 to 180 degrees, essentially creating out-of-distribution samples, we notice that the our uncertainty estimation clearly reflects this. This is a representation of unseen data, i.e. epistemic source of uncertainty in the third row in~\autoref{fig:motivating_example}. 

Our choice of experiments is guided by~\autoref{tab:overview}: we will evaluate our method CUSP \wrt input noise-specific experiments by way of adversarial attacks, to target noise using label flips, and to out-of-distribution detection (both in-domain and out-of-domain).


%% file: latex/40_method.tex
\section{Uncertainty Surrogates}
\label{sec:method}
%
%
The crux of our work is in forcing the pen-utlimate layer of a deep network to resemble a known pattern. This layer then serves as a proxy for a measure of prediction uncertainty, which we term \emph{uncertainty surrogate}. An Uncertainty surrogate can be trained using a visual pattern, in which case it can additionally provide qualitative feedback. 

%

\subsection{Problem setup}
We are considering the case of a deep classifier $\mathcal{F}$ as composed of a feature creating hidden part $h$ that computes feature vector $\mathbf{s} \in \mathbb{R}^m$, which is then used by the usual linear classifier $f$ to compute the prediction vector $\mathbf{y}$. That is, $\mathbf{s} = f(\mathbf{x})$ and $\mathbf{y} = f(\mathbf{s})$. The~\autoref{fig:schema} explains this schematically.

The network $\mathcal{F}$ is trained on observations from a dataset $\mathcal{D} \sim (X, T)$  where the inputs are random variables $X \in \mathbb{R}^d$ with $d > m$ commonly, and $T \in \mathcal{K} \sim \lbrace 1, 2 \dots k\rbrace$ a categorical variable representing at most $k$ classes. We train our model end-to-end on pairs of observations $\lbrace \mathbf{x_i}, \mathbf{t}_i \rbrace_{i =1}^N$ and simultaneously obtain features $\mathbf{s}$ and prediction $\mathbf{y}$.

It is commonly understood that the pen-ultimate layer features $\mathbf{s}$, give a high level representation of the data that is most informative for downstream tasks. In our work, we aim to take advantage of this property of $\mathbf{s}$ by additionally enforcing it to resemble a class-dependent visual pattern. To this end, we introduce class-specific patterns, $\lbrace  \mathbf{p}_k| k, \forall k \in \mathcal{K} \rbrace$, where $\mathbf{p_k}  \in \mathbb{R}^ m $ same as $\mathbf{s}$.

While training, we optimize over a sum of two losses, one for predicting the correct class label (classification) and the other, which forces the surrogate features to resemble the visual pattern (reconstruction). 
The loss function therefore takes the form:
\begin{align}
    \mathcal{L} \:\: &= \mathcal{L}_1 + \alpha*\mathcal{L}_2, \text{where}\\
    \mathcal{L}_1 & = -\sum_{i}^{\mathcal{K}}t_i\log(y_i),\\
    \mathcal{L}_2 &= -\sum_{i}^{m}p_{t_i}\log({s_i }) + (1-p_{t_i})\log(1-s_i),
\end{align}

\noindent with $\mathcal{L}_1$ being the categorical cross entropy (CCE) loss for classification into $C$ classes \wrt ground truth $\mathbf{t}$, and $\mathcal{L}_2$ being the logits-based binary cross entropy (BCE) loss for reconstruction of the $m$ pixels of the surrogate pattern $\mathbf{p}$. We choose $\alpha = 0.5$ for all our experiments. 

We use the binary cross entropy loss for training the network because for our binary patterns it provided better visual quality as compared to mean square error (MSE). Please note that the patterns need not be limited to a single layer binary or gray-scale ones, they can be in color, or be composed of multiple binary/gray-scale patterns, as long as the shape of the surrogate layer $\mathbf{s}$ can be adapted to match that of the surrogate pattern $\mathbf{p}$.

Apart from using an additional loss function, the only other modification done to the regular deep learning process to make the size of the pen-ultimate layer equal to the size of the surrogate pattern of choice. That is, if we use $16\times 16$ patterns, we set the size of the penultimate layer to $256$.

\input{./latex/41_fig_patterns.tex}

\subsection{Choice of surrogate patterns}
To train the uncertainty surrogates, the choice of the type for the surrogate patterns can provide different properties. In~\autoref{fig:ex_patterns}, we introduce an initial proposal of visual cues that we use as patterns. If the patterns are binary (or gray-scale) in pixel values, as is the case our work, the surrogate patterns for each class can be orthogonal to each other or non-orthogonal.

If orthogonal patterns are chosen, the network is encouraged to introduce orthogonality in the decision making earlier than at the output. Intuitively, and according to literature, orthogonality of activations improves the accuracy of deep classifiers~\cite{bansal2018can}. However, the orthogonal patterns may not carry any explicit visual meaning (see first row of~\autoref{fig:ex_patterns}). More importantly, the number of bright to dark pixels can drop significantly if the number of classes is large. This may introduce far too much sparsity (because the dark pixels in the patterns represent zeros) than may be acceptable for a reasonable performance. 

Non-orthogonal patterns can take any visual symbol and can have any desired degree of sparsity. They can even be error correcting codes like QR codes, which rely on Reed-Solomon error correction and do not necessarily have a visual meaning. It is also possible to express domain knowledge in the choice of surrogate patterns. For example, if two of the several classes in a classification scenario are too similar and hence difficult to separate, the surrogate patterns may be designed or chosen to ensure they are as dissimilar as possible according to some metric like euclidean distance. This will force the activations to be dissimilar making the classification task more discriminative for this.

\subsection{Obtaining prediction uncertainty}
\label{subsec:driver2}

Having trained a network with the surrogate patterns, at inference time, for a new sample $\mathbf{x}^*$, the deep model outputs $\mathbf{y}^*, \mathbf{s}^* = \mathcal{F}(\mathbf{x^*})$. We compute a scalar uncertainty score
\begin{equation}\label{eq:cusp_score}
	u  = \delta(\mathbf{p_{y*}}, \mathbf{s^*}),
\end{equation}

\noindent which is the distance between $\mathbf{x}^*$ and the surrogate pattern $\mathbf{p_{y*}}$ corresponding to the predicted label. For $\delta$ in~\autoref{eq:cusp_score}, one could choose either BCE, which was used during training or MSE to measure the distance, i.e the uncertainty between the surrogate pattern and the  reconstruction. We find MSE to be as effective or better than BCE as shown in the experiments section. 

The function $\delta$ need not be limited to MSE or BCE. Interestingly, since the reconstructed pattern is of a visual nature, we can use a light CNN to replace the $\delta$ function. This approach of using a secondary network for uncertainty prediction is superior to using the BCE or MSE reconstruction error as shown in~\autoref{fig:roc_curves} albeit at a slightly higher computational cost.

As for the complexity of CUSP, since at test time we obtain the surrogates at the same time as the label predictions, our method requires only computing element wise distance in the dimension of the feature layer, i.e. pattern shape $m$. Both the distance functions are quite efficient computationally, linear in the number of pixels of the surrogate pattern, with MSE having a slight edge over BCE.

%% file: latex/41_fig_patterns.tex
\renewcommand{\colwidth}{0.09\linewidth}
\renewcommand{\gap}{@{\hspace{0.5mm}}}

\begin{figure}
\centering
\begin{tabular}{c\gap c\gap c\gap c\gap c\gap c\gap c\gap c\gap c\gap c\gap}


\includegraphics[width=\colwidth]{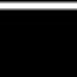}&
\includegraphics[width=\colwidth]{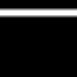}&
\includegraphics[width=\colwidth]{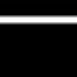}&
\includegraphics[width=\colwidth]{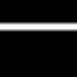}&
\includegraphics[width=\colwidth]{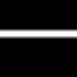}&
\includegraphics[width=\colwidth]{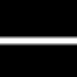}&
\includegraphics[width=\colwidth]{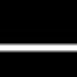}&
\includegraphics[width=\colwidth]{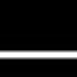}&
\includegraphics[width=\colwidth]{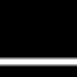}&
\includegraphics[width=\colwidth]{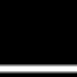}\\


\includegraphics[width=\colwidth]{./images/patterns_digits/00.pdf}&
\includegraphics[width=\colwidth]{./images/patterns_digits/01.pdf}&
\includegraphics[width=\colwidth]{./images/patterns_digits/02.pdf}&
\includegraphics[width=\colwidth]{./images/patterns_digits/03.pdf}&
\includegraphics[width=\colwidth]{./images/patterns_digits/04.pdf}&
\includegraphics[width=\colwidth]{./images/patterns_digits/05.pdf}&
\includegraphics[width=\colwidth]{./images/patterns_digits/06.pdf}&
\includegraphics[width=\colwidth]{./images/patterns_digits/07.pdf}&
\includegraphics[width=\colwidth]{./images/patterns_digits/08.pdf}&
\includegraphics[width=\colwidth]{./images/patterns_digits/09.pdf}\\

\includegraphics[width=\colwidth]{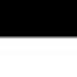}&
\includegraphics[width=\colwidth]{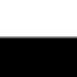}&
\includegraphics[width=\colwidth]{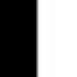}&
\includegraphics[width=\colwidth]{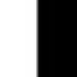}&
\includegraphics[width=\colwidth]{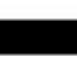}&
\includegraphics[width=\colwidth]{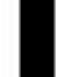}&
\includegraphics[width=\colwidth]{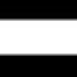}&
\includegraphics[width=\colwidth]{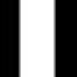}&
\includegraphics[width=\colwidth]{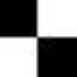}&
\includegraphics[width=\colwidth]{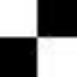}\\

\includegraphics[width=\colwidth]{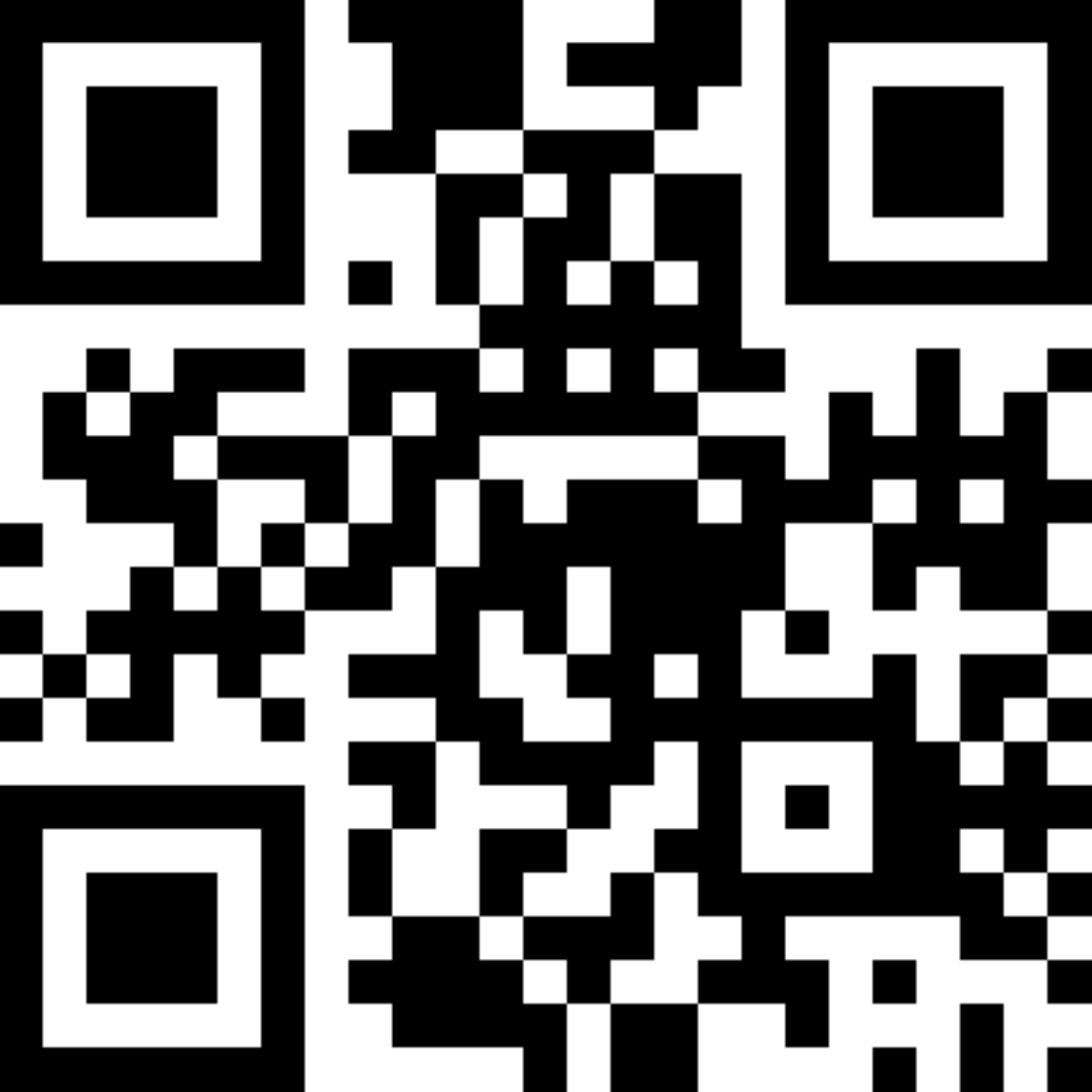}&
\includegraphics[width=\colwidth]{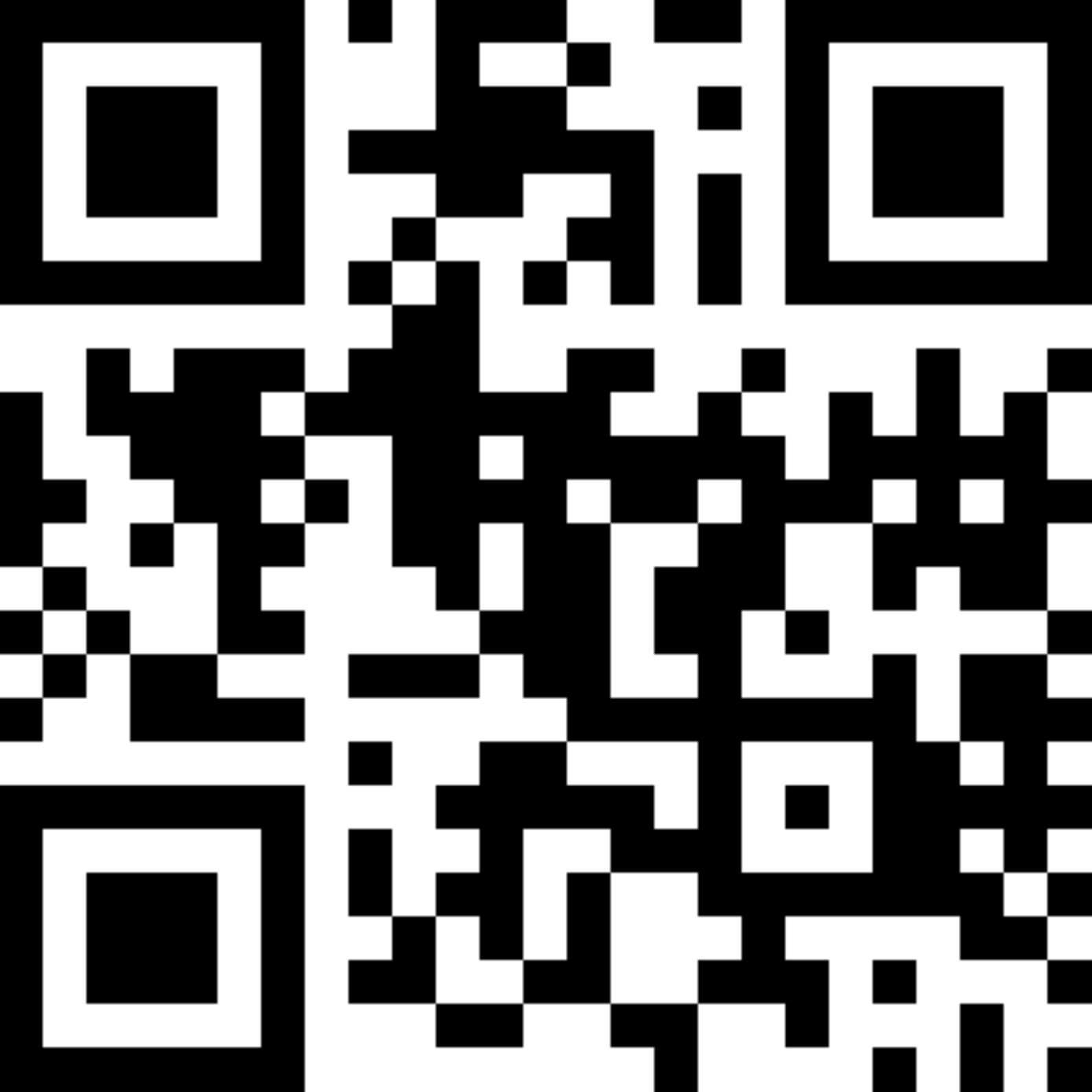}&
\includegraphics[width=\colwidth]{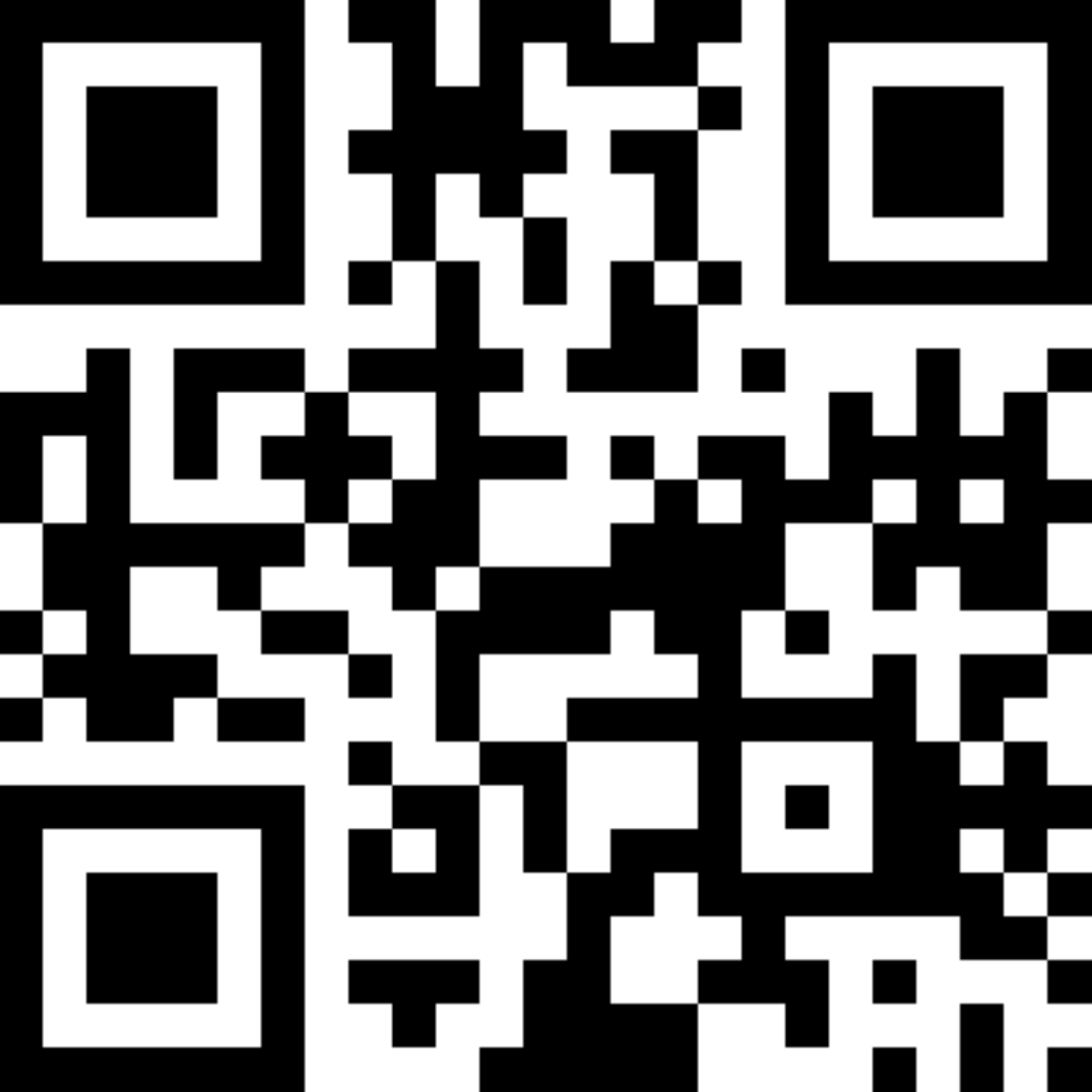}&
\includegraphics[width=\colwidth]{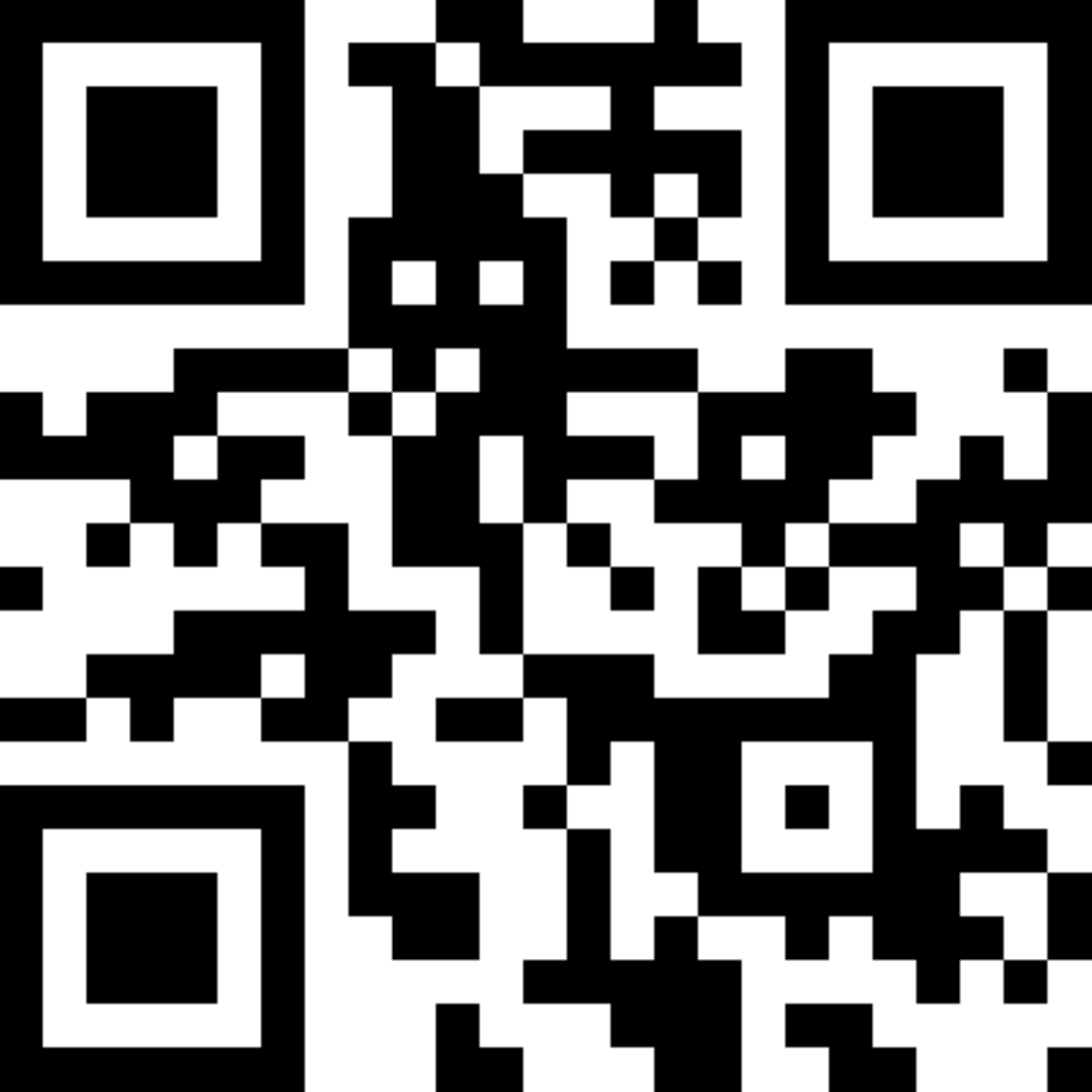}&
\includegraphics[width=\colwidth]{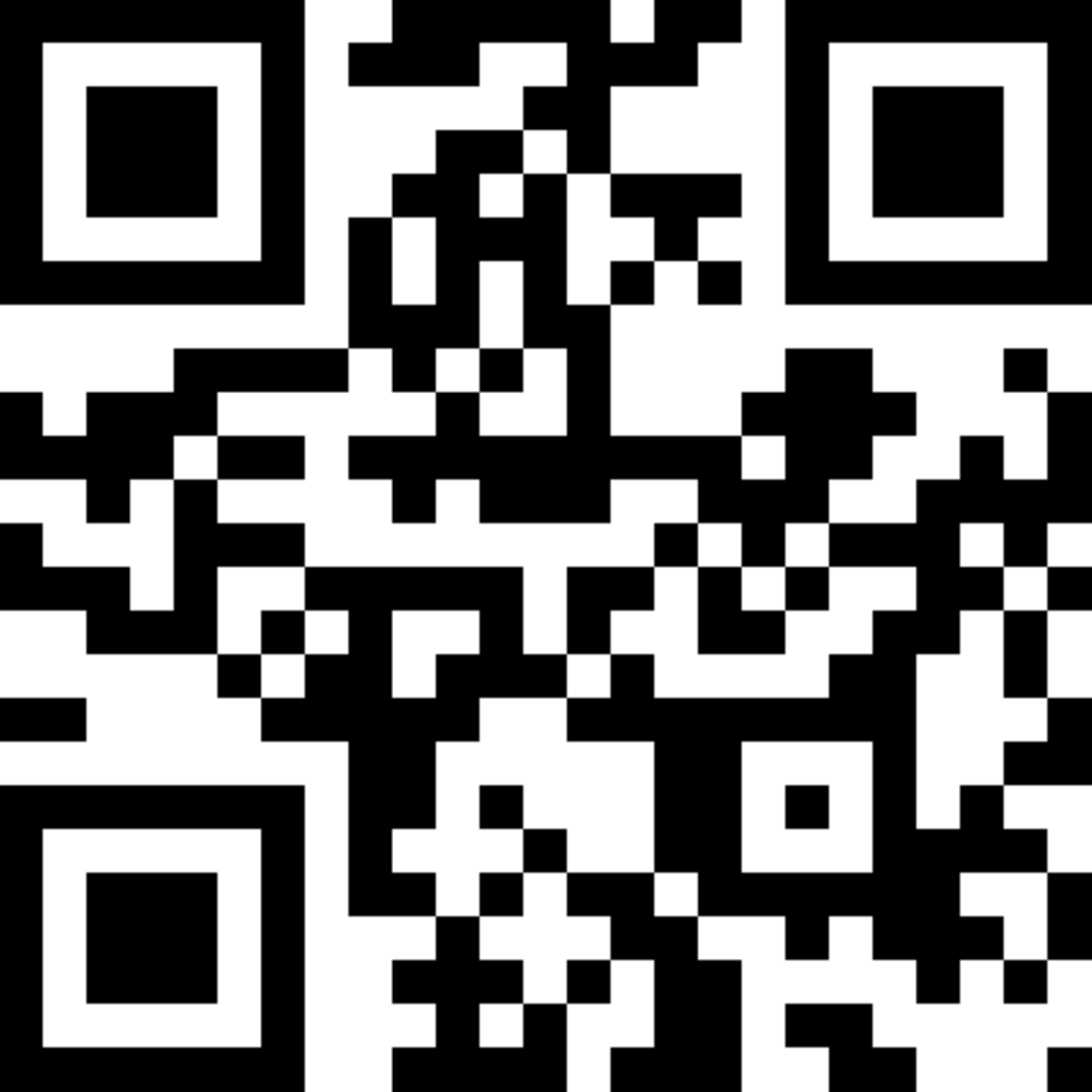}&
\includegraphics[width=\colwidth]{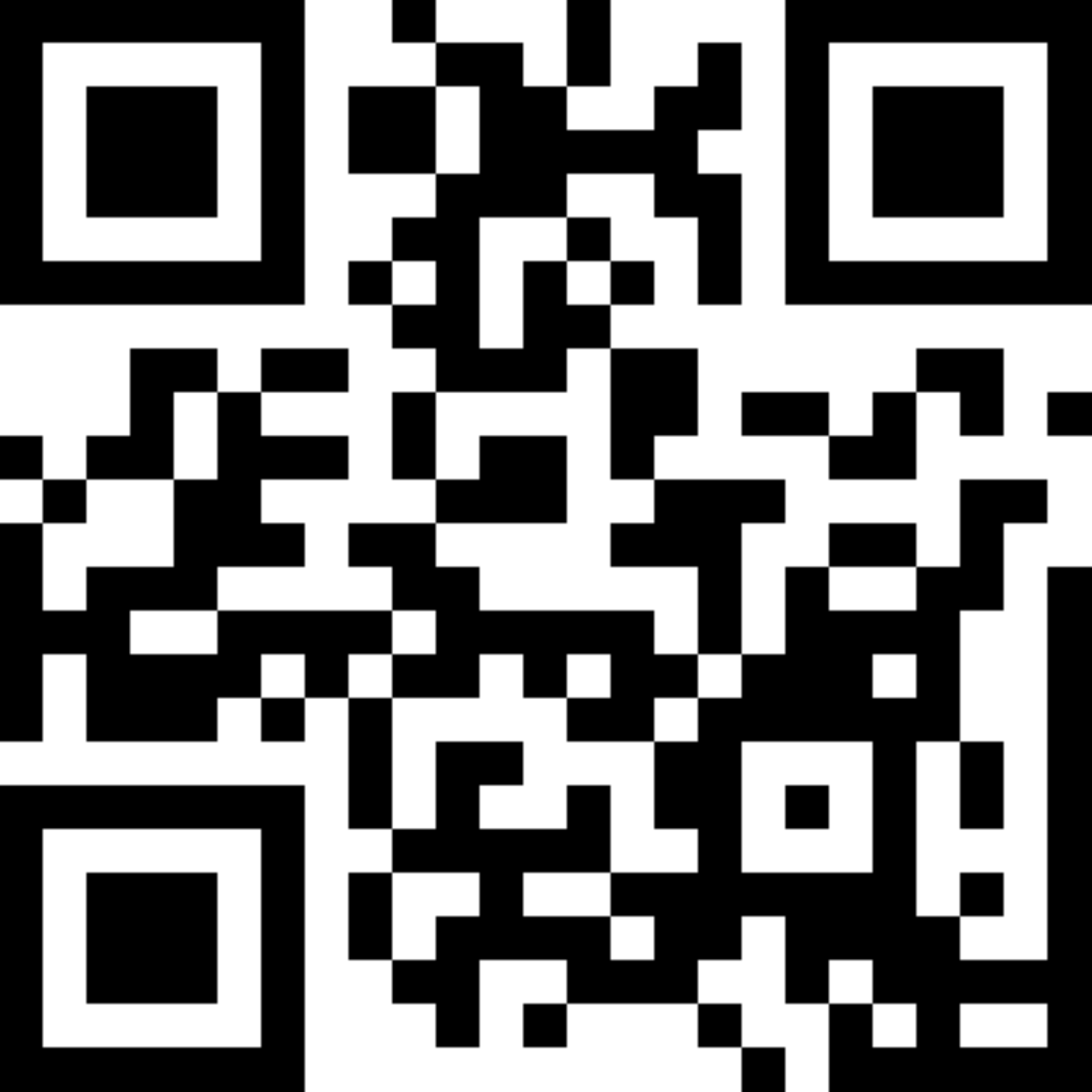}&
\includegraphics[width=\colwidth]{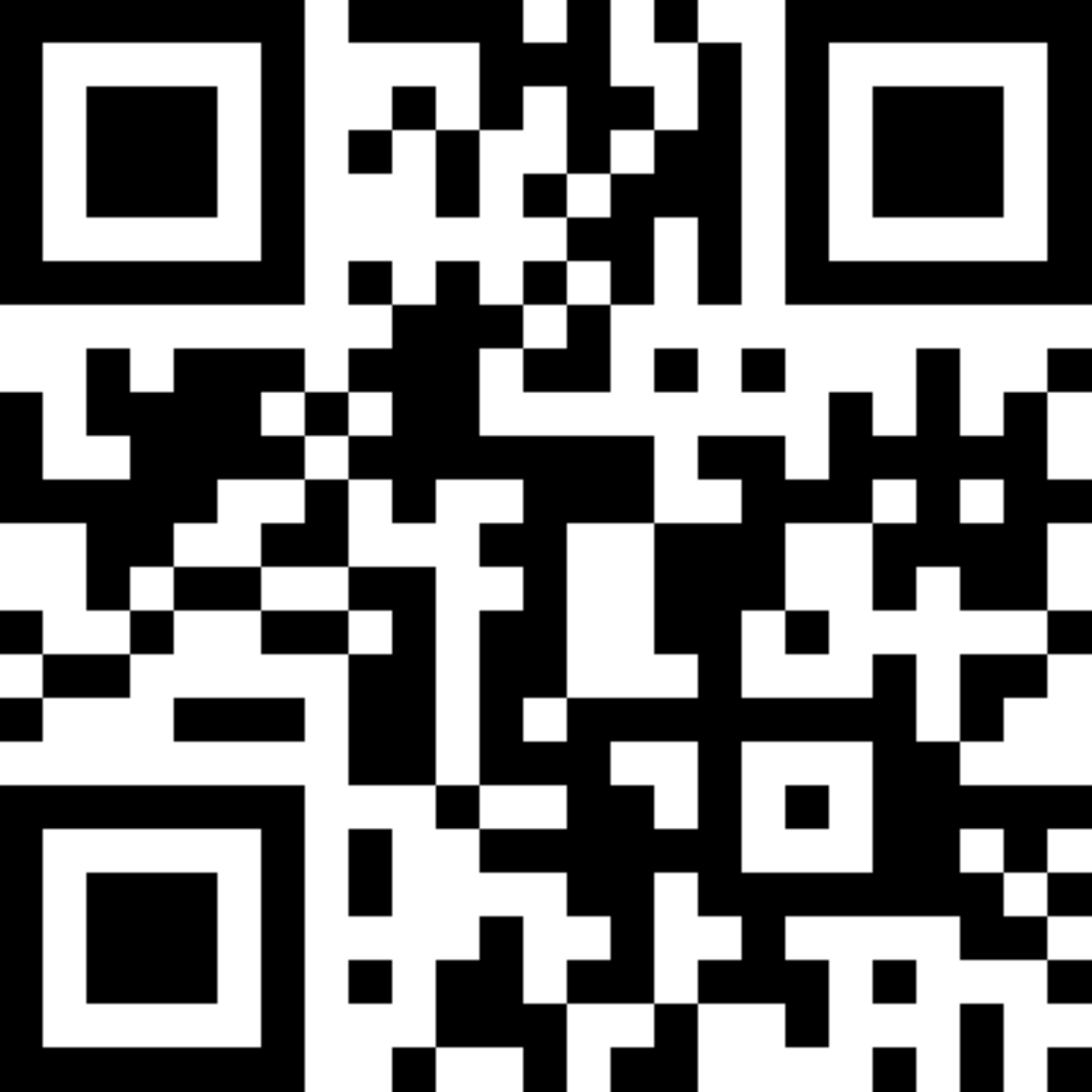}&
\includegraphics[width=\colwidth]{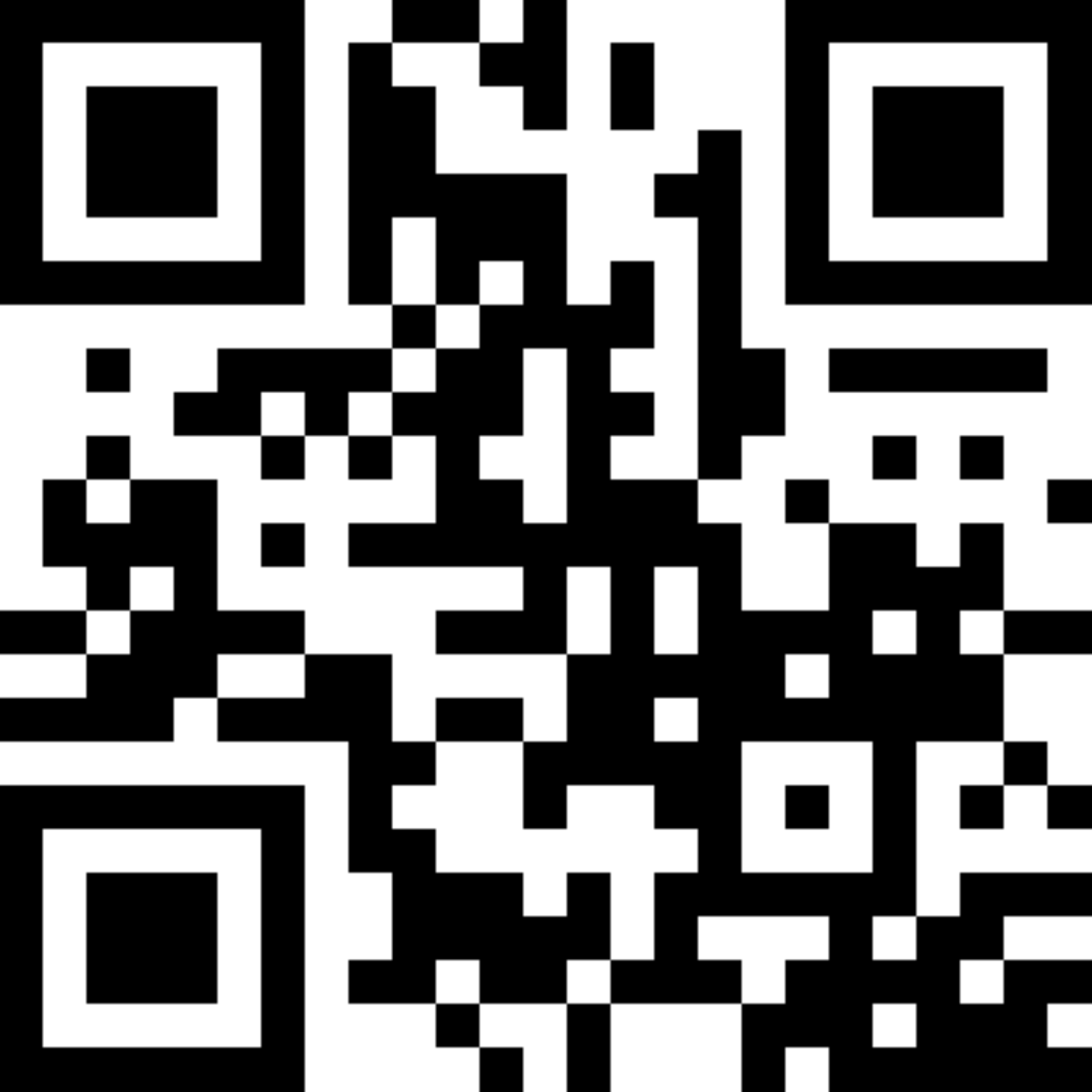}&
\includegraphics[width=\colwidth]{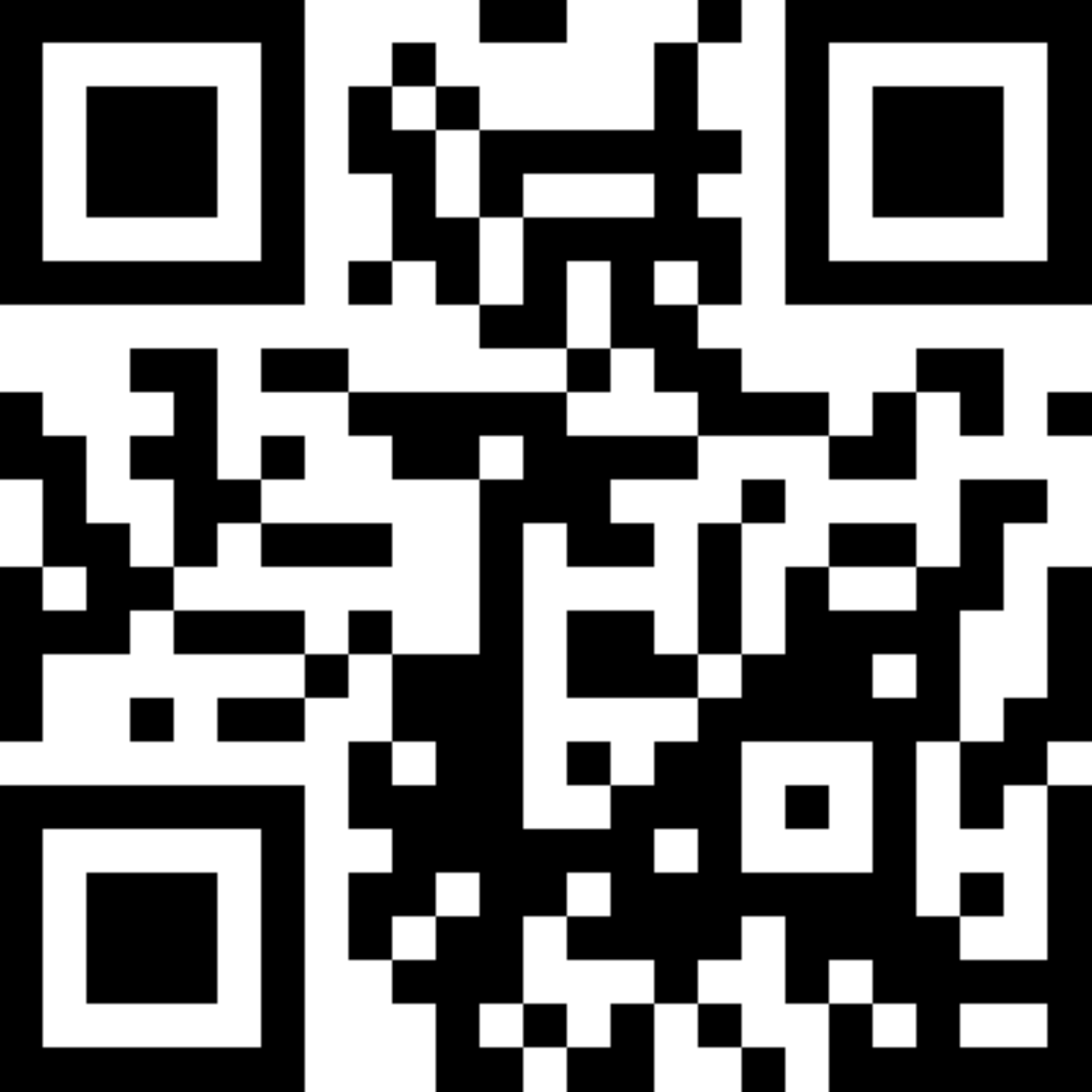}&
\includegraphics[width=\colwidth]{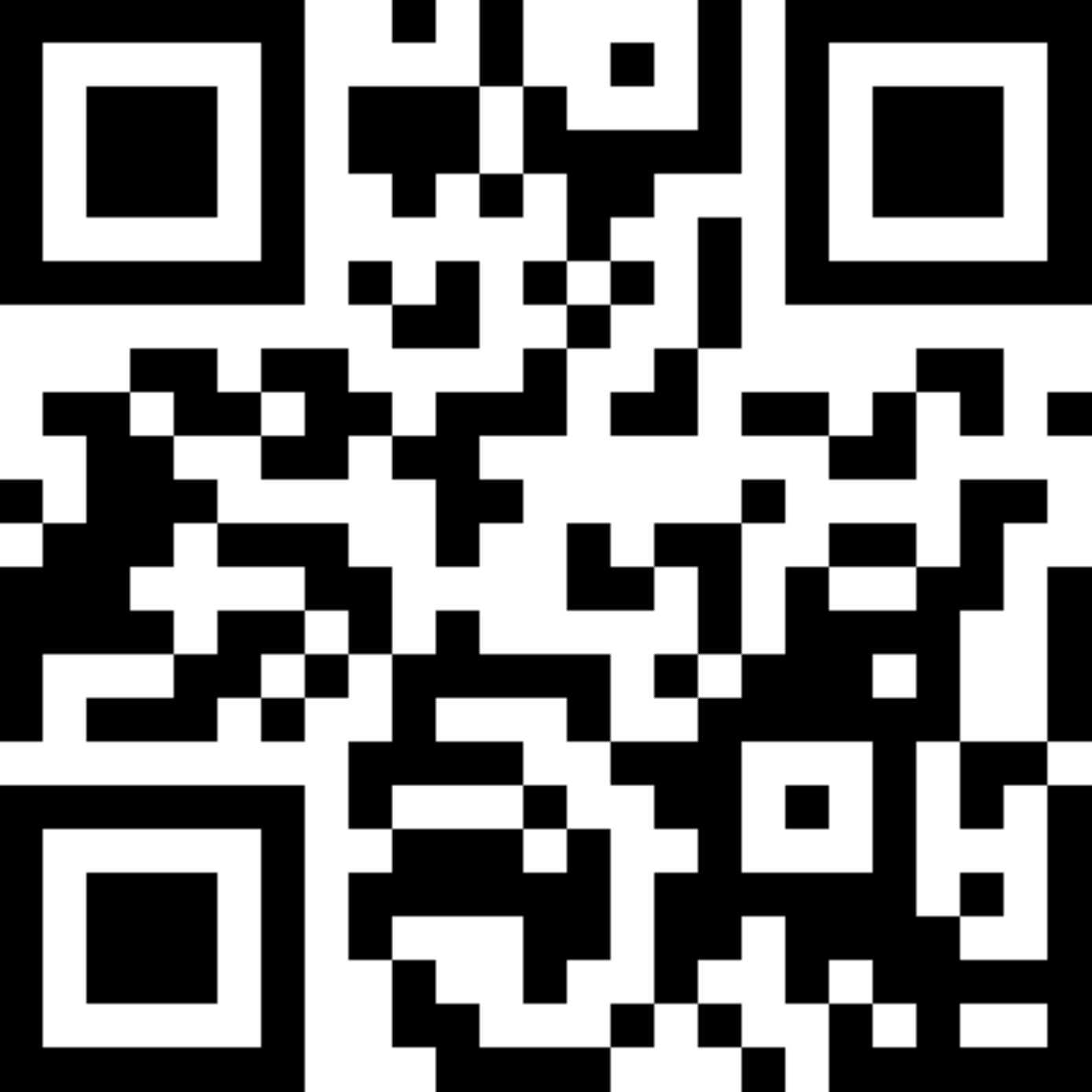}\\

\includegraphics[width=\colwidth]{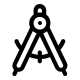}&
\includegraphics[width=\colwidth]{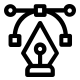}&
\includegraphics[width=\colwidth]{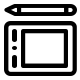}&
\includegraphics[width=\colwidth]{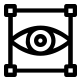}&
\includegraphics[width=\colwidth]{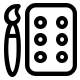}&
\includegraphics[width=\colwidth]{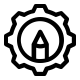}&
\includegraphics[width=\colwidth]{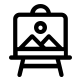}&
\includegraphics[width=\colwidth]{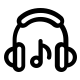}&
\includegraphics[width=\colwidth]{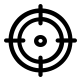}&
\includegraphics[width=\colwidth]{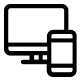}\\

\multicolumn{10}{c}{}\\

\end{tabular}
\caption{Examples of surrogate patterns, \ie fixed patterns, each associated with a distinct class. The bright and dark pixels represent ones and zeros, respectively. The top-row patterns are orthogonal. All others are non-orthogonal patterns. The patterns can even be error-correcting codes like QR-codes. Please note that such patterns can also be in color although we stick to binary ones in this paper.}
\label{fig:ex_patterns}
\end{figure}

%% file: latex/50_experiments.tex
\section{Experiments}
\label{sec:experiments}
\input{./latex/51_fig_prcurves.tex}



In what follows we introduce an extensive empirical evaluation for the benefits and various use cases of the surrogates uncertainties. In the coming sub-sections we organize our results that address the following questions in order:
\begin{enumerate}[noitemsep]

	\item Is CUSP applicable to cases where the source of uncertainty is epistemic? 
	\item Is CUSP applicable to cases where the source of uncertainty is aleatoric? 
	\item How reliable is the CUSP score?
	\item Does CUSP increase the robustness of a deep model?
	\item Does CUSP improve the interpretability of the model?
\end{enumerate}

\paragraph{Experimental setup} In all experiments we use the implementation of PreActResNet18~\cite{preact} with modification only to the last layer to adjust for the  dimensions of the surrogate patterns. 
We use the same hyperparameters and architecture for CUSP and all baselines with the only difference being the added reconstruction loss term to the CUSP case.

For CUSP we have three parameters the user can set: the distance  measure for $\delta$ in \autoref{eq:cusp_score}, the surrogate pattern and the size of the pattern and consequently the feature layer.

We pick the datasets of MNIST and fashionMNIST for grayscale images, and those of CIFAR10, SVHN and CIFAR100 for color images. All experiments were done on Tesla GPUs P100-PCIE.

\paragraph{Baselines}
We compare against multiple state-of-the-art uncertainty estimates that rely on the last or penultimate layer of a classifier. 
We follow a similar setup as the one of Tagasovska \etal~\cite{tagasovska2018single}. We include as baselines, Bayesian linear regression (covariance) baseline, distance to nearest training points (distance), largest softmax score \cite[largest]{hendrycks2016baseline}, absolute difference between the two largest softmax scores (functional), softmax entropy (entropy), geometrical margin \cite[geometrical]{wang2014new}, ODIN \cite{odin}, random network distillation \cite[distillation]{burda2018exploration}, principal component analysis (PCA),  a random baseline (random), and an oracle trained to separate the in- and out-of- domain examples.
As a sanity check we also add a random score and an ``oracle", which means we compute the score knowing the ground truth labels.

\paragraph{Metrics} As commonly done in such setups we use the \emph{classification accuracy} of a network as well as the ROC \emph{area under the curve} (AUC). 
Essentially, ROC is a probability curve and the area under it represents a measure of class separability. The higher the AUC, the better the model is at predicting 0s as 0s and 1s as 1s.

\subsection{Epistemic uncertainty}
To confirm the applicability of CUSP under an epistemic sources of uncertainty we propose out-of-distribution experiments. We address the two cases of 1) same domain out-of-distribution detection and 2) different domain out-of-distribution detection. 
In both cases we are interested in the AUC score, which should give us an estimate on how different are the scores for samples seen during training as opposed to novel ones. 

\textbf{Different domain out-of-distribution detection}
We consider one full dataset as in-distribution and another one as out-of-distribution. In particular we consider the following cases: training on CIFAR10, while testing on SVHN, and vice versa; and training on MNIST, while testing on fashionMNIST, and vice-versa. 
From~ \autoref{tab:10vs10} we notice that CUSP has the best results on MNIST and SVHN. Moreover, we also note that it has consistently high scores across all datasets, which is not the case with the rest of the baselines. 

\paragraph{Same domain out-of-distribution detection}
Here we split each of the four datasets at random into five ``in-domain'' classes and five ``out-of-domain'' classes.
We note that this experimental setup is much more challenging than  the previous one since in this case the in and out of distribution examples are more similar than if taken from completely different datasets.

From~\autoref{tab:5vs5} we notice that CUSP has the highest AUC scores across all datasets in this more challenging setup and with a very small variance across different configurations and datasets.

\begin{table*}
	\begin{center}
		\resizebox{0.68\textwidth}{!}{
			\begin{tabular}{lrrrrrrrrr}
				\toprule
				& \textbf{CIFAR10} & \textbf{fashion} & \textbf{MNIST} & \textbf{SVHN} & \textbf{Average}  \\
				\midrule
				\textbf{certificates} & $0.92 \pm 0.01$ & $0.90 \pm 0.05$ & $0.68 \pm 0.10$ & $0.81 \pm 0.00$ & 0.83 $\pm$ 0.09\\
				\textbf{covariance}  & $0.63 \pm 0.00$ & $0.81 \pm 0.00$ & $0.99 \pm 0.00$ & $0.58 \pm 0.00$ & 0.75 $\pm$ 0.16\\
				\textbf{distillation} & $0.74 \pm 0.10$ & $0.81 \pm 0.02$ & $0.96 \pm 0.02$ & $0.55 \pm 0.01$ & 0.76 $\pm$ 0.15 \\
				\textbf{entropy}      & \textbf{0.96 $\pm$ 0.02} & \textbf{0.95 $\pm$ 0.01} & $0.77 \pm 0.03$ & 0.95 $\pm$ 0.00 & $0.91 \pm 0.08$ \\
				\textbf{functional}   & $0.94 \pm 0.01$ & $0.93 \pm 0.01$ & $0.79 \pm 0.01$ & $0.94 \pm 0.00$ &$ 0.90 \pm 0.06$ \\
				\textbf{geometrical}  & $0.87 \pm 0.10$ & $0.72 \pm 0.07$ & \textbf{0.98 $\pm$ 0.01} & $0.79 \pm 0.15$ & $0.84 \pm 0.09$\\
				\textbf{largest}      & $0.96 \pm 0.02$ & $0.92 \pm 0.02$ & $0.75 \pm 0.03$ & \textbf{$0.95 \pm 0.00$} & $0.89 \pm 0.08$\\
				\textbf{odin}         & $0.93 \pm 0.07$ & $0.95 \pm 0.01$ & $0.77 \pm 0.04$ & $0.94 \pm 0.01$ & $0.90 \pm 0.07$\\
				\textbf{pca}          & $0.63 \pm 0.16$ & $0.61 \pm 0.06$ & $0.66 \pm 0.10$ & $0.52 \pm 0.01$ & $0.60 \pm 0.05$\\
				\textbf{random}         & $0.51 \pm 0.00$ & $0.51 \pm 0.01$ & $0.50 \pm 0.0$ & $0.51 \pm 0.01$ &$ 0.50 \pm 0.00 $\\
				\textbf{oracle}         & $0.94 \pm 0.00$ & $0.99 \pm 0.01$ & $0.99 \pm 0.00$ & $0.99 \pm 0.01$ & $0.99 \pm 0.00$ \\
				\bottomrule
				\hline
				\hline
				CUSP digits & 0.80443& 0.91854 & 0.94160 &0.92626 & $0.90 \pm 0.05$\\
				CUSP non ortho & 0.89647 & 0.80717& 0.95507& 0.95507 & $0.90 \pm 0.06$\\
				CUSP ortho  &0.85852& 0.90069& 0.98671&0.96140& $0.93 \pm 0.05$\\
				\textbf{CUSP BCE} &$0.85 \pm 0.04$ & $0.87 \pm 0.04$ & $0.96 \pm 0.01$&$ 0.93 \pm 0.05$ &$0.90  \pm 0.04$\\
				\bottomrule
				CUSP digits  & 0.85528& 0.80395 &  0.97382& 0.95556 & $0.90 \pm 0.07$\\
				CUSP non ortho  & 0.92223 & 0.93617&  0.96616& 0.95641 &$ 0.94 \pm 0.01$\\
				CUSP ortho   &0.90031 & 0.94077 & 0.93507 &0.96002 & $0.93 \pm 0.02$\\
				\textbf{CUSP MSE }&$0.89\pm 0.03 $& $0.89 \pm 0.03$ & $0.96 \pm 0.01$ & \textbf{0.96 $\pm$ 0.01} & \textbf{0.92 $\pm$ 0.03}\\
				\bottomrule
				&&&&&\\
			\end{tabular}\label{tab:10vs10}
		}
		\caption{AUC scores for out-of-distribution uncertainty. By column: average and standard deviation across different hyper-parameters. By row: average and standard deviation  per method across datasets.}
		\label{tab:10vs10}
	\end{center}
\end{table*}

\begin{table*}
	\begin{center}
		\resizebox{0.68\textwidth}{!}{
			\begin{tabular}{lrrrrrrrrrr}
				\toprule
				& \textbf{CIFAR10} & \textbf{fashion}  & \textbf{MNIST}  & \textbf{SVHN}  &   \textbf{Average} \\
				\midrule
				\textbf{certificates}& $0.75 \pm 0.01$ & $0.74 \pm 0.00$ & $0.87 \pm 0.01$ & $0.87 \pm 0.03$ & $0.80 \pm 0.06$ \\
				\textbf{covariance}   & $0.54 \pm 0.00$ & $0.70 \pm 0.08$ & $0.88 \pm 0.09$ & $0.62 \pm 0.00$ &$0.69 \pm 0.12$ \\
				\textbf{distance}     & $0.62 \pm 0.03$ & $0.79 \pm 0.00$ & $0.87 \pm 0.09$ & $0.67  \pm 0.02$ &$ 0.74 \pm 0.10 $\\
				\textbf{distillation} & $0.56 \pm 0.03$ & $0.68 \pm 0.04$ & $0.72 \pm 0.13$ & $0.79 \pm 0.03$ & $0.69 \pm 0.08 $\\
				\textbf{entropy}      & $0.73 \pm 0.00$ & $0.81 \pm 0.00$ & $0.93 \pm 0.02$ & $0.92 \pm 0.01$ &$ 0.85 \pm 0.08$\\
				\textbf{functional}   & $0.72 \pm 0.00$ & $0.83 \pm 0.01$ & $0.94 \pm 0.01$ & $0.92 \pm 0.01$ & $0.85 \pm 0.08 $\\
				\textbf{geometrical}  & $0.70 \pm 0.08$ & $0.67 \pm 0.10$ & $0.80 \pm 0.11$ & $0.80 \pm 0.09$ &$ 0.74 \pm 0.05$\\
				\textbf{largest}      & $0.69 \pm 0.03$ & $0.82 \pm 0.00$ & $0.89 \pm 0.04$ & $0.92 \pm 0.01$ &$ 0.83 \pm 0.08$\\
				\textbf{odin }        & $0.71 \pm 0.03$ & $0.77 \pm 0.05$ & $0.89 \pm 0.01$ & $0.92 \pm 0.02$ & $0.82 \pm 0.08 $\\
				\textbf{pca}          & $0.57 \pm 0.06$ & $0.56 \pm 0.05$ & $0.54 \pm 0.05$ & $0.58 \pm 0.05$ & $0.56 \pm 0.14 $\\
				\textbf{random}          & $0.50 \pm 0.00$ & $0.51 \pm 0.00$ & $0.50 \pm 0.0$ & $0.52 \pm 0.01$ & $0.50 \pm 0.00$ \\
				\textbf{oracle }         & $0.94 \pm 0.00$ & $1.00 \pm 0.01$ & $0.99 \pm 0.00$ & $0.99 \pm 0.01$ &$ 0.99 \pm 0.00$ \\
				\bottomrule
				\hline
				\hline
				CUSP ortho        &  0.78775 &  0.88854 & 0.96838 &  0.93540 & $0.90 \pm 0.06$  \\
				CUSP non ortho & 0.73533 &0.86589 &  0.96487 & 0.92602 & $0.87 \pm 0.09$\\
				CUSP digits & 0.79648 &0.84193&0.94251&0.94348 & $0.88 \pm 0.06$\\
				\textbf{CUSP BCE} & $0.77 \pm 0.02 $& \textbf{0.87 $\pm$ 0.02} & \textbf{0.96 $\pm$ 0.01} & \textbf{0.93 $\pm$ 0.01}  & \textbf{0.88 $\pm$ 0.02}\\
				\bottomrule
				
				CUSP digits & 0.75920& 0.74232& 0.90327 &  0.87009 &$ 0.82 \pm 0.07$\\
				CUSP non ortho & 0.79259&0.78787 & 0.95465 &0.92378 & $0.86 \pm 0.07$\\
				CUSP ortho  & 0.79780&  0.89320 &0.95070 & 0.89052 &$ 0.88 \pm 0.05$\\
				\textbf{CUSP  MSE}& \textbf{0.78 $\pm$  0.02} &$0.81 \pm 0.06$&$0.94 \pm 0.02$&$0.90 \pm 0.02 $ & $0.86 \pm 0.06$\\
				\bottomrule
				&&&&&\\
			\end{tabular}
		}
		\caption{AUC scores for same-domain uncertainty. By column: average and standard deviation across different hyper-parameters. By row: average and standard deviation  per method across datasets.}
		\label{tab:5vs5}
	\end{center}
\end{table*}

\subsection{Aleatoric uncertainty}
\begin{figure}
	\centering
	\includegraphics[width=1.0\columnwidth]{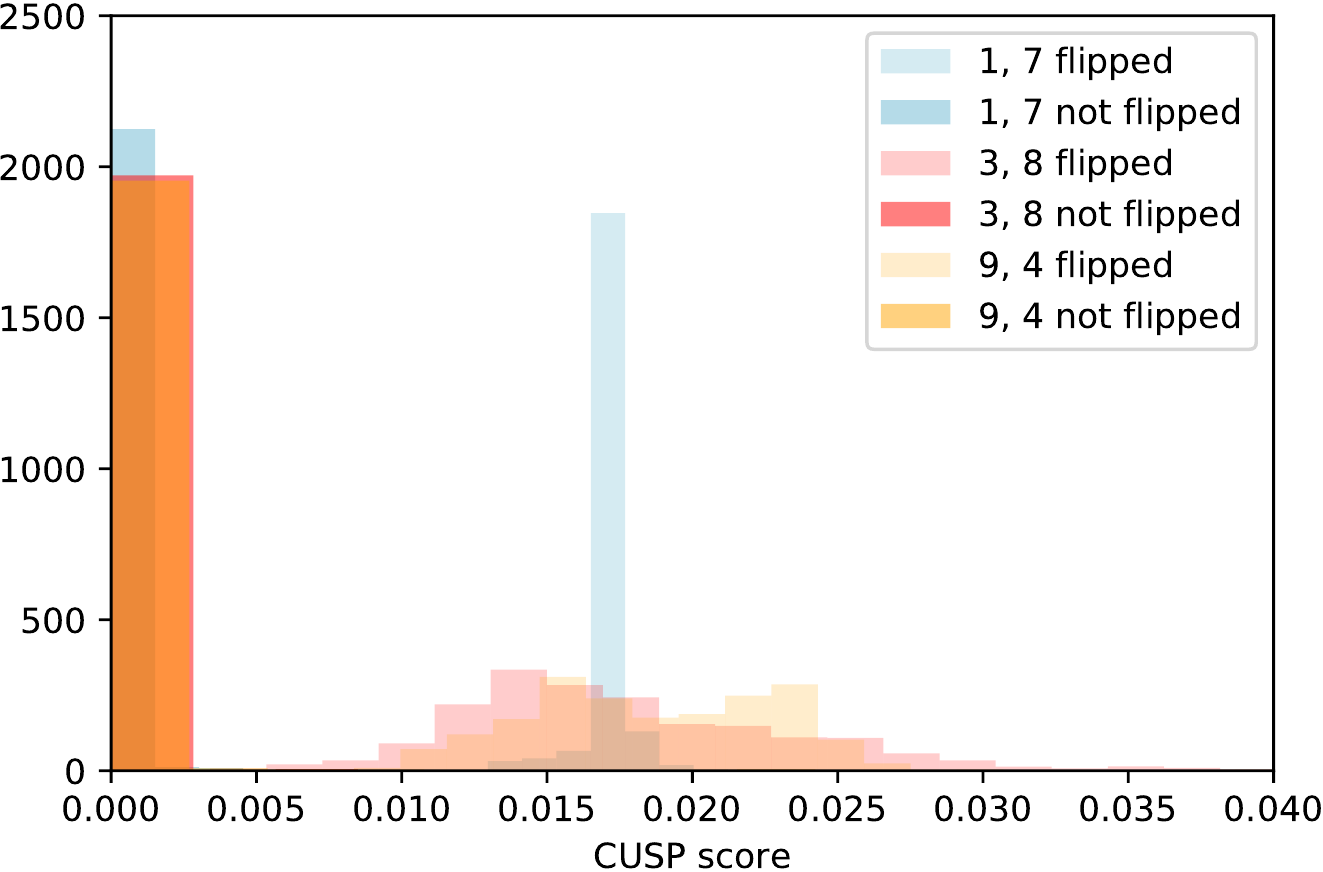}
	\caption{Sensitivity of CUSP scores to heteroskedastic noise. Uncertainty is higher when trained on flipped cases. Experiment done with orthogonal patterns and MSE loss.}
	\label{fig:flipped}
\end{figure}

To investigate the reliability of CUSP under noise uncertainty in the targets, we propose the following ``flipped class" experiment. We chose pairs of visually similar classes, and for 30\% of the training samples we flip the target labels. More specifically, some 1s get labeled as 7, some 4s as 9 and some 3s as 8. In \autoref{fig:flipped} we observe that CUSP is indeed sensitive to heteroscedastic noise, \ie, as desired, in the case of flipped labels the uncertainty scores are consistently higher.

\subsection{Reliability of CUSP }
In this section we evaluate the sensitivity of the surrogates with regards to different choice of parameters. 


	

\begin{figure*}
	\includegraphics[width=1.0\textwidth]{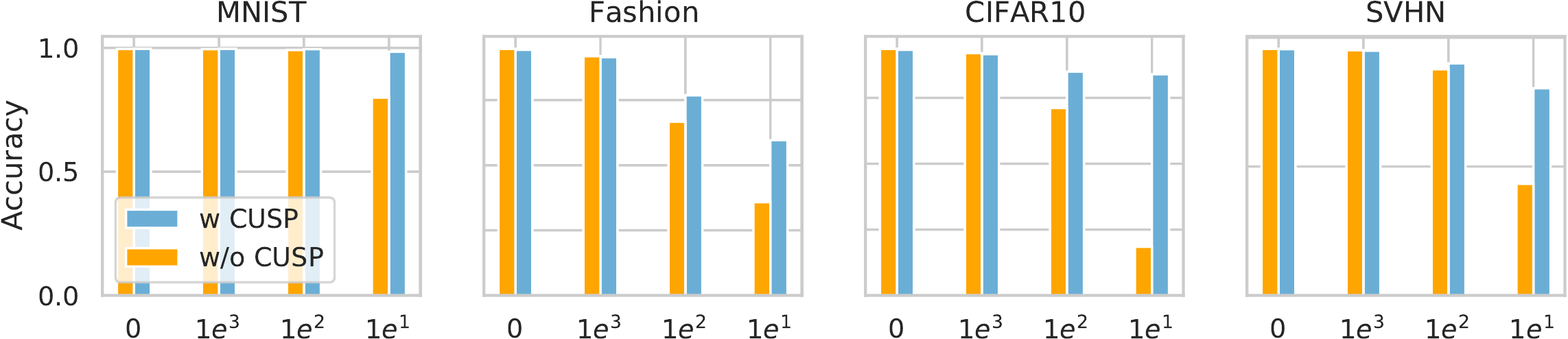}
	\caption{Accuracy of classifiers under adversarial attacks. Error bars come from different $\epsilon$ values for the FGM. It can be noticed that at any level of adversarial attacks, the  network trained with CUSP remains robust. }
	\label{fig:adversarial_results}
\end{figure*}

The setup is as follows: we split each of the datasets into three parts, in the ratio 10:1:1. The first part is used to train a PreActResNet18 model to predict the given patterns and the class labels. The second part is used to predict the patterns, classes, and the MSE based uncertainties using the trained PreActResNet18. The predicted patterns are now assigned binary labels of \emph{true} or \emph{false} depending on whether the PreActResNet18 predicts the ground truth labels correctly or not. Given these binary labels as targets, the predicted patterns corresponding original patterns, and the predicted MSE-based uncertainty scores are used to train a light 3-layer CNN to predict the binary label with a single sigmoid output and BCE loss. The complement of this scalar output is then treated as the uncertainty prediction. The third part of the dataset is used to test the secondary CNN. As a note, since the number of correct predictions largely exceeds incorrect ones, we use focal loss~\cite{lin2017focal} for training with $\gamma=2$.

\autoref{fig:roc_curves} shows the comparison of BCE, MSE, and the secondary CNN using an ROC curve. To plot the curve, we threshold each of the three uncertainties at a 100 thresholds ranging from 0 to 1. Each threshold splits the given patterns into \emph{true} and \emph{false} predictions, which are then compared with the binary ground truth labels we collected on the second part of the dataset. This permits us to compute counts of true positives, false positives, true negatives, and false negatives at each threshold value of uncertainty.
The curves show that in all three datasets the CUSP scores whether computed with BCE, MSE, or the second classifer (CNN proba) consistently outperforms the strongest baselines.

\subsection{CUSP robustness  check }
Adversarial attacks could be considered as a form of an aleatoric noise in the inputs, but if the produced example is actually outside of the training manifold, this becomes an epistemic source of uncertainty as well. In this experiment, we are interested in actually, how sensitive will a CUSP network be if exposed to a white-box attack.
To this end we evaluate the accuracy of the classifier when a network has been trained with the surrogate loss as opposed to standard training.
We introduce different levels of additive adversarial noise following the Fast Gradient Method, (FGM)~\cite{goodfellow2014explaining}, and at each level we measure the impact on the classifier accuracy. 

The gradient sign method uses the gradient of the underlying model to produce adversarial examples. An  image $x$ is manipulated by adding or subtracting a small error  $\epsilon$
to each pixel. Adding errors in the direction of the gradient means that the image is intentionally altered so that the model classification fails.
The following formula describes the fast gradient sign method:
\begin{equation*}
	x^\prime= x + \epsilon \text{sign}(\nabla_x J(\theta, x, y))
\end{equation*}
$\nabla_x J(x, y)$ is the gradient of the models' loss function with respect to the original input pixel vector $x$, $y$ is the true label vector for $x$ and  $\theta$ is the model parameter vector.


From \autoref{fig:adversarial_results}, we notice that that when confronted with adversarial attacks, the networks with surrogate regularization manage to consistently keep higher accuracy as compared to their baseline counterparts. This confirms the increased robustness due to the CUSP training.


\begin{table}[]
	
	\resizebox{0.5\textwidth}{!}{
	\begin{tabular}{lrrr|rrr}
		\toprule
		 & \textbf{entropy} & \textbf{largest} & \textbf{functional} & \textbf{BCE} & \textbf{MSE} & \textbf{CNN}\\
		\midrule
		\textbf{Fashion}  & 0.8100 & 0.8216 & 0.8210 & 0.9693 & 0.9922 & 0.9927\\
		\textbf{CIFAR-10}  & 0.7017 & 0.7613 & 0.7585 & 0.8841 & 0.9239 & 0.9325\\
		\textbf{CIFAR100}  & 0.8023 & 0.8304 & 0.8220 & 0.9986 & 0.9985 & 0.9813\\
		\bottomrule
		& & & & & & \\
	\end{tabular}
	}
	\caption{AUC for the RoC curves of ~\autoref{fig:roc_curves}. Compared to the state-of-the-art, the uncertainty predicted by our approach is consistently better \wrt AUC scores.}
	\label{tab:auc}
\end{table}



%% file: latex/51_fig_prcurves.tex




\begin{figure*}
    \centering
    \begin{tabular}{ccc}

\includegraphics[width=0.32\textwidth]{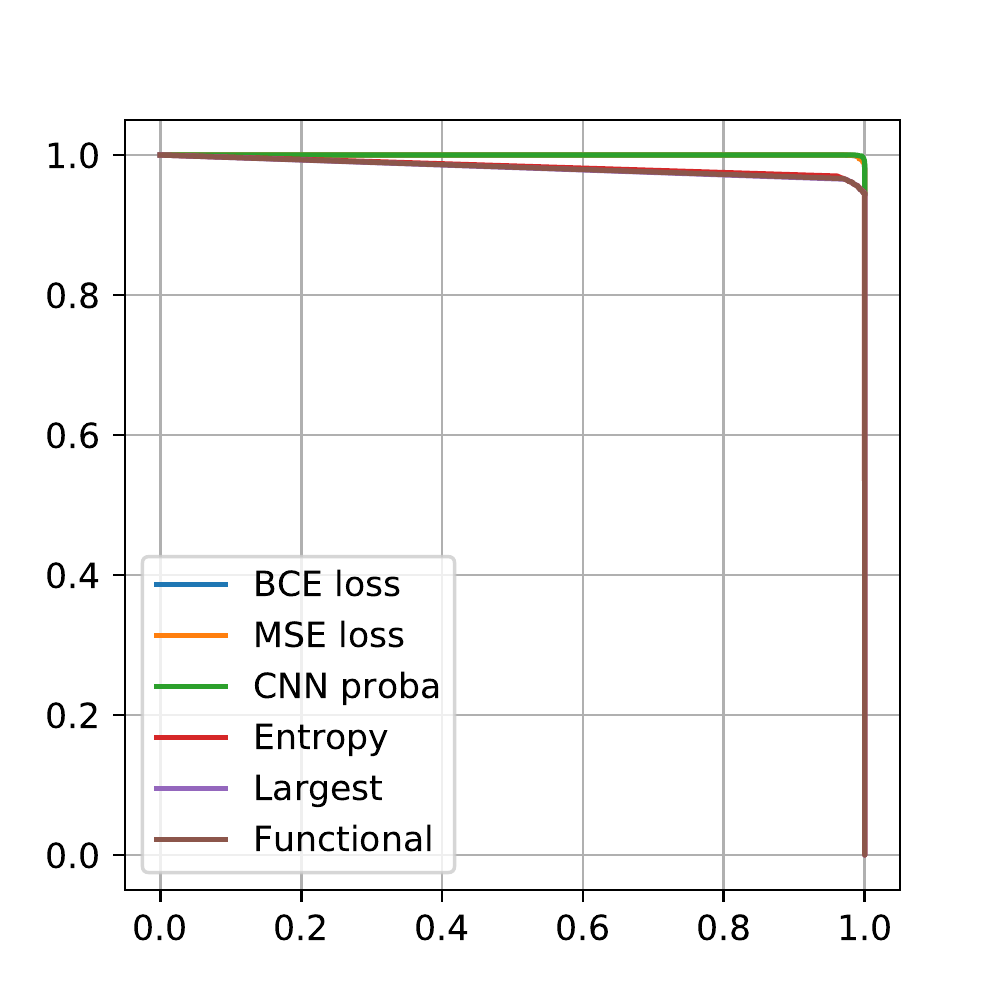}&
\includegraphics[width=0.32\textwidth]{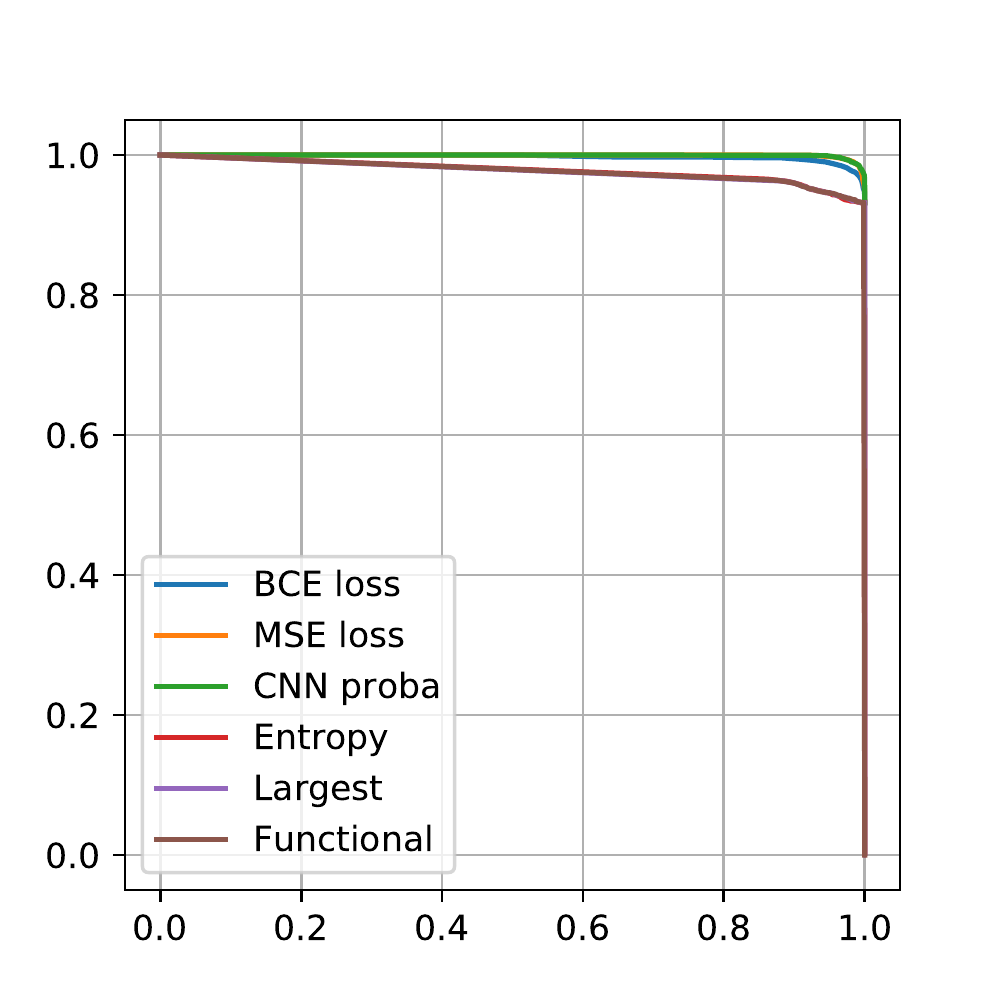}&
\includegraphics[width=0.32\textwidth]{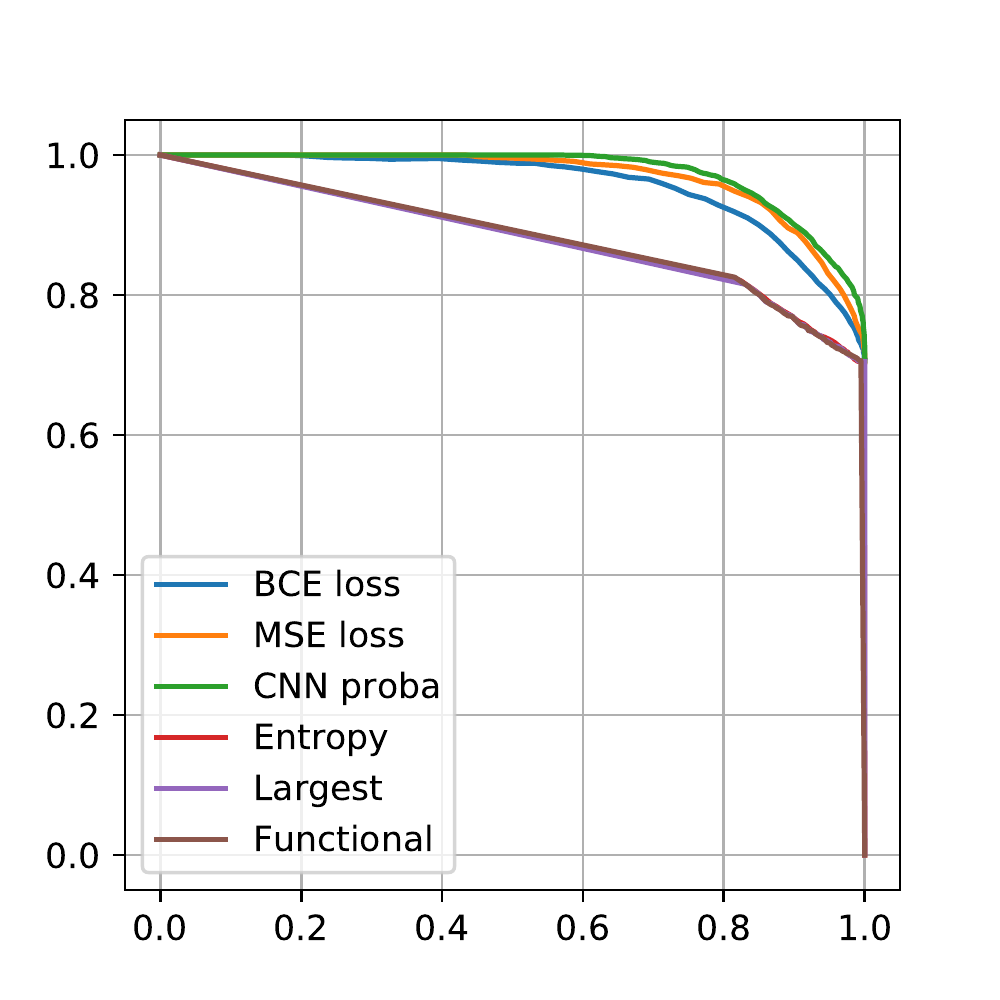}\\

        \includegraphics[width=0.32\textwidth]{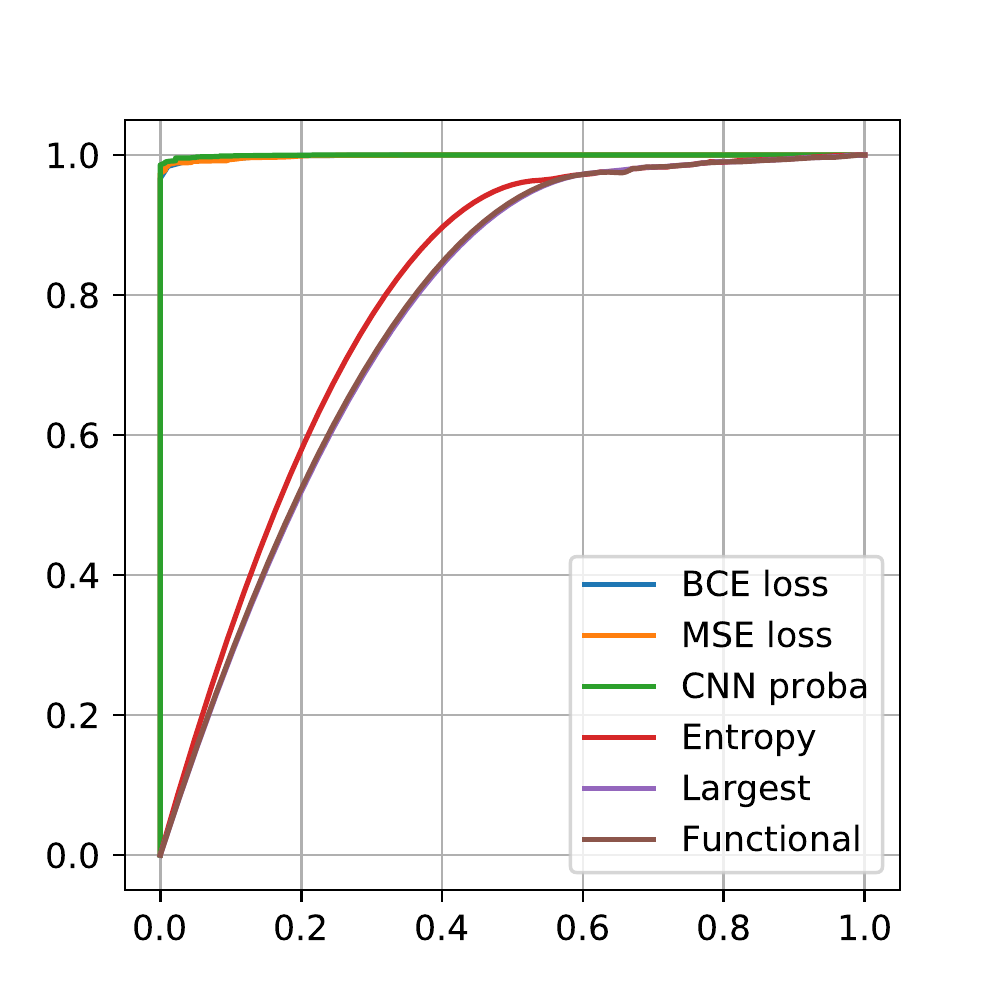}&
        \includegraphics[width=0.32\textwidth]{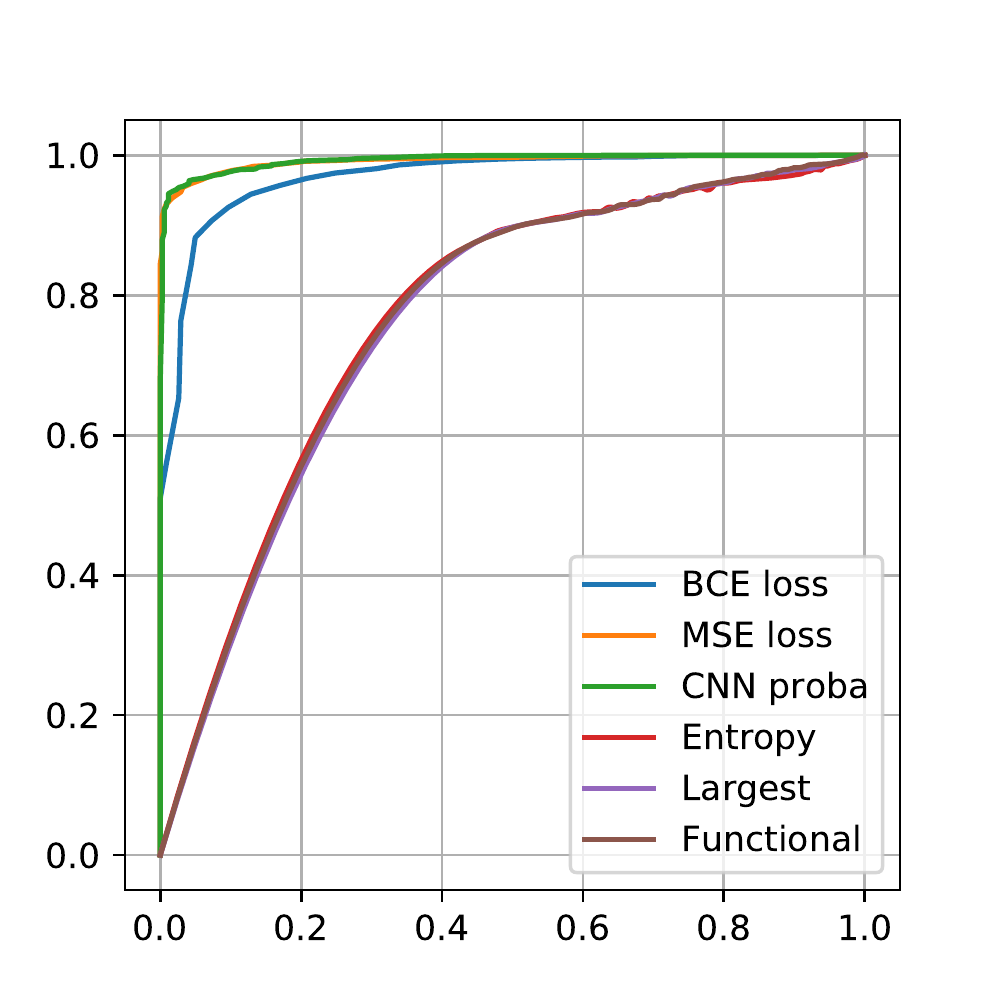}&
        \includegraphics[width=0.32\textwidth]{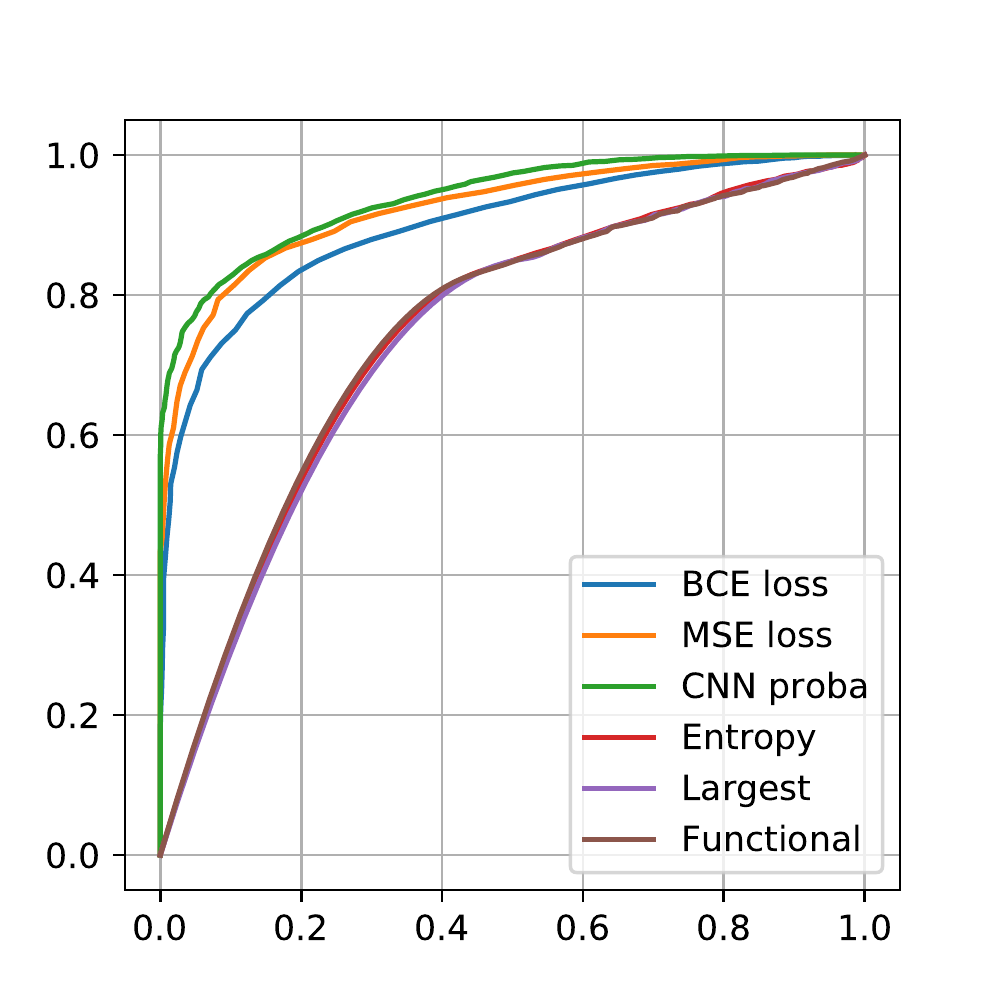}\\
    
    \end{tabular}
    \caption{Plots of precision versus recall (top row) and ROC (bottom row) to compare the uncertainty estimate provided by BCE, MSE, and the secondary CNN trained on the recreated pattern. The three plots from left to right are for the datasets of FashionMNIST, CIFAR-10, CIFAR-100, respectively. The plots show that BCE, MSE, and our secondary CNN peform better than competing methods.}
    \label{fig:roc_curves}
\end{figure*}

%% file: latex/60_conclusion.tex
\section{Conclusion}
\label{sec:conclusion}
Providing the uncertainty of prediction for deep learning, which is mandatory for any critical application of machine learning, was difficult or reserved to computationally expensive methods. In this paper, we introduce a simple method to make a regular deep network provide the prediction and the confidence of its prediction, better than existing methods as shown qualitatively and quantitatively. To achieve this, we place a simple constraint on the readily available features of the penultimate layer to resemble a pre-determined visual pattern. Not only does this provide better uncertainty estimates in a computationally efficient manner, but it also allows us to detect out-of-distribution examples and to resist adversarial attacks at no extra cost. Additionally, the reconstructions of the predetermined patterns provide visual feedback during training as well as testing, which, along with the uncertainty estimates, renders the training process and the predictions of the network more interpretable.
 Finally, we also hope that the visual interpretability of our uncertainty score will open new directions for communicating the behavior and results of deep models to the end users.

